\documentclass[10pt,twocolumn,letterpaper]{article}

\usepackage[pagenumbers]{cvpr} 

\usepackage[dvipsnames]{xcolor}

\usepackage{graphicx}
\usepackage{amsmath}
\usepackage{amssymb}
\usepackage{booktabs}

\usepackage{cite}
\usepackage{amsfonts}
\usepackage{algorithmic}
\usepackage{textcomp}
\usepackage{xcolor}
\usepackage{makecell}
\usepackage{array,hhline}
\usepackage{comment}
\usepackage{color}
\usepackage{overpic}
\usepackage{multirow}
\usepackage{tabularx}
\usepackage{adjustbox}
\usepackage{times}
\usepackage{epsfig}
\usepackage{float}

\definecolor{cvprblue}{rgb}{0.21,0.49,0.74}
\usepackage[pagebackref,breaklinks,colorlinks,citecolor=cvprblue]{hyperref}

\def\paperID{3052} 
\def\confName{CVPR}
\def\confYear{2024}

\title{Toward Accurate and Temporally Consistent Video Restoration from Raw Data}

\author{Shi Guo\textsuperscript{1} \quad Jianqi Ma\textsuperscript{1,2} \quad Xi Yang\textsuperscript{1,2} \quad Zhengqiang Zhang\textsuperscript{1,2} \quad Lei Zhang\textsuperscript{1,2} \\
\textsuperscript{1}The Hong Kong Polytechnic University; \textsuperscript{2}OPPO Research Institute\\
{\tt\small \{shiguo.guo, jianqi.ma, xxxxi.yang, zhengqiang.zhang\}@connect.polyu.hk,} \\
{\tt\small cslzhang@comp.polyu.edu.hk}}

\begin{document}
\maketitle
\begin{abstract}
Denoising and demosaicking are two fundamental steps in reconstructing a clean full-color video from raw data, while  performing video denoising and demosaicking jointly, namely VJDD, could lead to better video restoration performance than performing them separately. In addition to restoration accuracy, another key challenge to VJDD lies in the temporal consistency of consecutive frames. This issue exacerbates when perceptual regularization terms are introduced to enhance video perceptual quality. To address these challenges, we present a new VJDD framework by consistent and accurate latent space propagation, which leverages the estimation of previous frames as prior knowledge to ensure consistent recovery of the current frame. A data temporal consistency (DTC) loss and a relational perception consistency (RPC) loss are accordingly designed. Compared with the commonly used flow-based losses, the proposed losses can circumvent the error accumulation problem caused by inaccurate flow estimation and effectively handle intensity changes in videos, improving much the temporal consistency of output videos while preserving texture details. Extensive experiments demonstrate the leading VJDD performance of our method in term of restoration accuracy, perceptual quality and temporal consistency. Codes and dataset are available at \url{https://github.com/GuoShi28/VJDD}.
\end{abstract}    
\section{Introduction}
\label{sec:intro}

In consumer grade imaging devices, denoising and demosaicking are two critical steps to reconstruct high-quality RGB images/videos from the sensor RAW data. Color demosaicking aims to reproduce the missing color components from the color-filter array (CFA) data captured by a single-chip CCD/CMOS sensor, while denoising aims to remove the noise caused by random photon arrival and readout circuitry. Many previous methods for demosaicking~\cite{ehret2019study,yang2019efficient,yan2019cross,liu2020new,jin2020review} and denoising~\cite{luisier2010image,zhang2017beyond,zhang2018ffdnet,plotz2018neural,liu2018non,guo2019toward,khademi2021self} in camera image signal processing (ISP) treat these two tasks sequentially. As a result, the denoising artifacts can be magnified in the demosaicking process, or the demosaicking errors can complicate the denoising process. 

Considering the correlations between the two tasks, jointly performing denoising and demosaicking (JDD)~\cite{condat2012joint,heide2014flexisp,gharbi2016deep,henz2018deep,kokkinos2019iterative,liu2020joint,guo2021joint,guo2022differentiable} has garnered considerable research attention. Gharbi \etal ~\cite{gharbi2016deep} explored a learning-based scheme for JDD, leading to superior results to traditional non-learning based JDD techniques~\cite{cok1987signal,heide2014flexisp}. Guo \etal~\cite{guo2021joint} demonstrated that the joint execution of denoising and demosaicking can yield better restoration results than performing these two tasks separately. Meanwhile, more advanced networks have been proposed for JDD, \eg, two-stage networks~\cite{qian2019trinity}, auto-encoder architectures~\cite{henz2018deep}, iterative structures~\cite{kokkinos2019iterative}, two-stage alignment framework~\cite{guo2022differentiable}. However, these JDD methods are primarily designed for single images (denoted by SJDD)~\cite{condat2012joint,heide2014flexisp,gharbi2016deep,henz2018deep,kokkinos2019iterative,liu2020joint} or burst images (denoted by BJDD)~\cite{guo2021joint,guo2022differentiable}, neglecting the scenarios of video restoration from raw data. With the increasing demand for high-quality videos, it is imperative to develop new video restoration algorithms in raw domain.  In this paper, we aim to address this challenging issue and present an effective framework of joint denoising and demosaicking from video raw data (denoted by VJDD).

Compared to SJDD and BJDD, where the restoration accuracy is the key demand, VJDD has another key requirement, \ie, the temporal consistency among consecutive frames. This is because even each frame of the video can be well restored, there can be many structural and textural differences between adjacent frames, resulting in the unpleasant flickering artifacts when play the frames as a video. In addition, to mitigate the over-smoothing problem and enhance the perceptual quality of videos, many video restorations methods \cite{chan2022investigating,wu2022animesr,dai2022video,yang2021real,xie2023mitigating,jeelani2023expanding} introduce the perceptual loss \cite{johnson2016perceptual} and adversarial loss \cite{goodfellow2014generative} in the training process, which may generate more perceptually realistic details in each frame but exacerbate the temporal inconsistency between frames. To achieve temporal consistency, previous video processing algorithms, such as video colorization~\cite{lai2018learning,lei2019fully}, video style transfer~\cite{lai2018learning,wang2020consistent} and video generation~\cite{park2019preserving}, typically employ flow-based temporal regularization to align adjacent frames and constrain the differences between aligned frames. However, this approach may not work well for video restoration due to the accumulated errors by optical flow estimation and its tendency to smooth the aligned frames~\cite{dai2022video}. 

To address the above mentioned problems, we present an accurate and temporally consistent VJDD method. Our approach improves the design of network architecture as well as the loss functions. In terms of network architecture, we propose a consistent latent space propagation scheme by leveraging the deep prior of previous frames to regularize the recovery of current frame. In terms of temporal loss design, we propose a Data Temporal Consistency (DTC) loss and a Relational Perception Consistency (RPC) loss. By simulating a sequence of consecutive frames through random geometric transformations, the DTC loss employs the ground-truth transformations to ensure the consistency between frames. Compared with flow-based methods, DTC loss can avoid the error accumulation problem. Additionally, we apply both long-term and short-term DTC losses to model the temporal relations at different time scales. Inspired by the relation loss of \cite{dai2022video}, we design the RPC loss to encourage the perceptual similarity between the predicted video and the ground-truth. Unlike \cite{dai2022video}, which utilizes blur kernels to model pixel statistics in region-level, we model the relation by calculating the difference in perceptual feature space to avoid the over-smoothing problem. We conduct extensive experiments to assess the performance of our method in terms of restoration accuracy, perceptual quality and temporal consistency. The results demonstrate the significant improvements achieved by our method across all these aspects.
\section{Related work}
\label{sec:related_work}
\subsection{Joint Denoising and Demosaicking}
As two fundamental and correlated tasks in the image signal processing (ISP) pipeline of single chip CMOS/CCD sensors, image denoising and demosaicking have been extensively studied for many years. Early methods usually treat denoising~\cite{luisier2010image,zhang2017beyond,zhang2018ffdnet,plotz2018neural,liu2018non,guo2019toward,khademi2021self} and demosaicking~\cite{ehret2019study, yang2019efficient, yan2019cross, liu2020new, jin2020review} as separate and sequential steps in ISP. However, separately performing denoising and demosaicking can lead to accumulated errors during image restoration. Therefore, researchers~\cite{gharbi2016deep,qian2019trinity,henz2018deep,kokkinos2019iterative,ehret2019joint,liu2020joint,guo2021joint} later started to investigate joint denoising and demosaicking (JDD) methods. Gharbi \emph{et al.}~\cite{gharbi2016deep} demonstrated that using more challenging patches for training can effectively reduce the moir\'{e} artifacts in JDD. To obtain better restoration performance, several advanced JDD network structures have been developed, \eg, two-stage networks~\cite{qian2019trinity}, auto-encoder architectures~\cite{henz2018deep} and iterative structures~\cite{kokkinos2019iterative}. In addition to the JDD for a single image,  studies~\cite{guo2021joint,guo2022differentiable} have been conducted to perform the JDD task for burst images with low signal-to-noise ratio. Based on the fact that the green channel has twice the sampling rate of the red/blue channels and exhibits a higher signal-to-noise ratio, GCP-Net~\cite{guo2021joint} utilizes the green channel to guide feature learning and frame alignment. The 2StageAlign~\cite{guo2022differentiable} method employs a two-stage alignment framework to compensate for the large shifts between frames. 

Despite the many JDD methods for a single image and burst images, little work has been done on the JDD of videos, which requires not only the high restoration accuracy of each frame but also the temporal consistency among consecutive frames. We investigate this problem and propose an effective framework of video JDD in this paper.

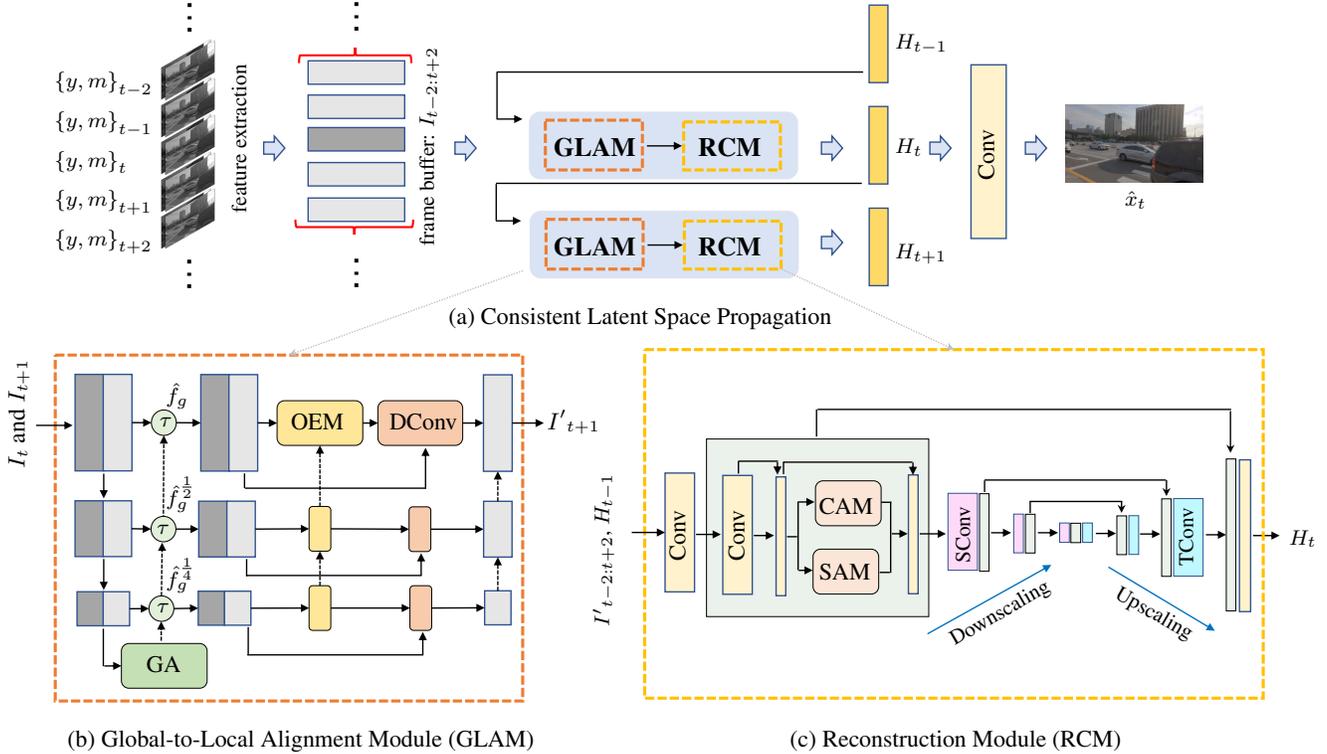
\begin{figure*}[!tbp]
\centering
\begin{overpic}[width=1.0\textwidth]{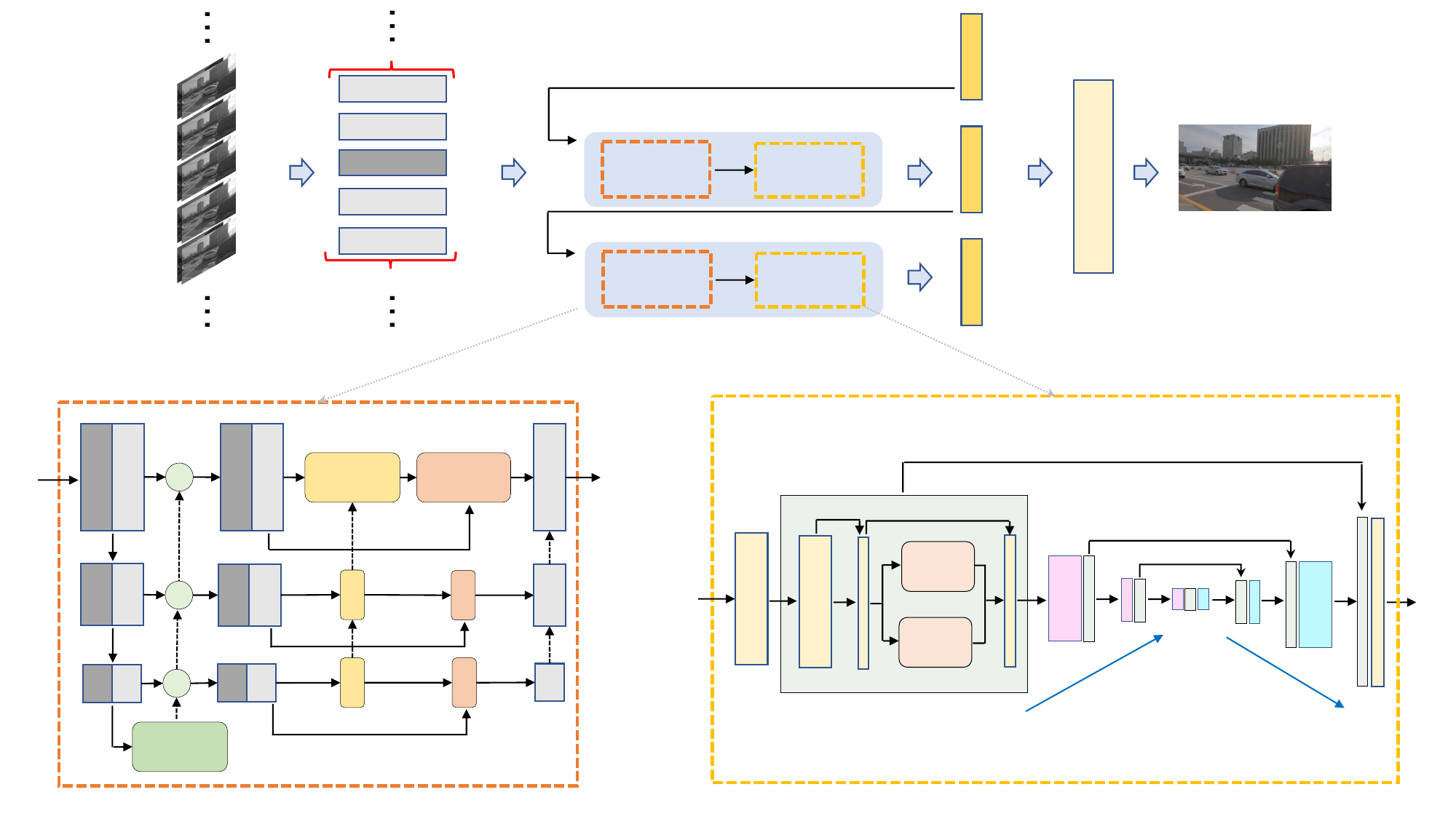}
    \put(4.0,49){\footnotesize ${\{y,m\}}_{t-2}$}
    \put(4.0,46){\footnotesize ${\{y,m\}}_{t-1}$}
    \put(4.0,43){\footnotesize ${\{y,m\}}_t$}
    \put(4.0,40){\footnotesize ${\{y,m\}}_{t+1}$}
    \put(4.0,37){\footnotesize ${\{y,m\}}_{t+2}$}

    \put(17.5,39.5){\footnotesize \rotatebox{90}{feature extraction}}
    \put(31.8,37){\footnotesize \rotatebox{90}{frame buffer: $I_{t-2:t+2}$}}

    \put(42,43.8){\normalsize \textbf{GLAM}}
    \put(53,43.8){\normalsize \textbf{RCM}}
    \put(42,36.2){\normalsize \textbf{GLAM}}
    \put(53,36.2){\normalsize \textbf{RCM}}
    
    \put(68,52){\footnotesize $H_{t-1}$}
    \put(68,44){\footnotesize $H_{t}$}
    \put(68,36){\footnotesize $H_{t+1}$}
    \put(74.3,42){\small \rotatebox{90}{Conv}}
    \put(85.5,40){\footnotesize $\hat{x}_t$}
    \put(34,31){\small (a) Consistent Latent Space Propagation}

    \put(5,-1.2){\small (b)  Global-to-Local Alignment Module (GLAM)}
    \put(45,8){\footnotesize \rotatebox{90}{${I'}_{t-2:t+2}$, $H_{t-1}$}}
    \put(60,-1.2){\small (c) Reconstruction Module (RCM)}

    \put(0.5,20){\small \rotatebox{90}{$I_t$ and $I_{t+1}$}}
    \put(41.5,23){\small ${I'}_{t+1}$}

    \put(11,4.5){\small GA}
    \put(12.5,11.0){\footnotesize $\hat{f}_{g}^{\frac{1}{4}}$}
    \put(12.5,17.2){\footnotesize $\hat{f}_{g}^{\frac{1}{2}}$}
    \put(12.5,25){\footnotesize $\hat{f}_{g}$}
    \put(11.8,9.0){\footnotesize $\tau$}
    \put(11.8,15.0){\footnotesize $\tau$}
    \put(12.0,23.2){\footnotesize $\tau$}
    \put(22.0,23.0){\small OEM}
    \put(29.5,23.0){\small DConv}

    \put(50.8,13){\small \rotatebox{90}{Conv}}
    \put(55.3,13){\small \rotatebox{90}{Conv}}
    \put(62.2,16.5){\small CAM}
    \put(62.2,11.5){\small SAM}
    \put(72.6,12.8){\footnotesize \rotatebox{90}{SConv}}
    \put(89.6,12.6){\footnotesize \rotatebox{90}{TConv}}
    \put(98,14.2){\footnotesize $H_t$}
    \put(72,6.5){\footnotesize \rotatebox{30}{Downscaling}}
    \put(84.5,10.2){\footnotesize \rotatebox{-30}{Upscaling}}

\end{overpic}\vspace{0.2cm}
\caption{(a) Illustration of our proposed consistent latent space propagation framework for VJDD. The extracted features are aligned by the global-to-local alignment module (GLAM), and then fed into the reconstruction module (RCM) to produce the clean full color video. (b) and (c) show the detailed structures of GLAM and RCM modules, respectively. In GLAM, GA, OEM and DConv represent the global alignment, offset estimation and deformable convolution modules, respectively. In RCM, CAM and SAM refer to the channel attention and spatial attention modules, respectively.}
\label{fig:Overall}
\vspace{-0.2cm}
\end{figure*}

\subsection{Temporal Consistency}
Existing methods for achieving temporal consistency in video processing primarily focus on the design of consistency regularization terms in network training~\cite{lei2019fully,park2019preserving,sajjadi2018frame,wang2018video,zhang2021learning,song2022tempformer,dai2022video}. The most widely adopted regularization technique is to align the adjacent frames using optical flow and enforce the spatially aligned pixels to be close in intensity. Such a flow-based approach has been successfully employed in video colorization~\cite{lai2018learning,lei2019fully}, video style transfer~\cite{lai2018learning,wang2020consistent} and video generation~\cite{park2019preserving}. However, in applications of video restoration, the accuracy of optical flow estimation will be largely affected by the noise and video intensity variations. As discussed in \cite{dai2022video}, flow-based regularization often leads to over-smoothed videos. Recently, Dai \emph{et al.} \cite{dai2022video} proposed a region-level relation loss, which compares the statistics changes of predicted frames with those of the ground-truth frames. However, this approach uses a blur function to compute the statistical property, which neglects image textures. Song \emph{et al.} \cite{song2022tempformer} generated overlapped frames and applied temporal constraints on them to ensure consistency. Although this method can well model the temporal consistency, it incurs additional computational costs due to the repeated generation of video frames. In this work, we introduce a consistent latent space propagation approach with a data temporal consistency loss and a relational perceptual consistency loss. Our method can accurately model temporal consistency without sacrificing video details.

\section{Method}

\subsection{Overall Framework}
The objective of video joint denoising and demosiacking (VJDD) is to reconstruct a clean RGB video $\mathcal{X} = \{ x_t \}_{t=1}^{N}$ from the mosaicked noisy raw video $\mathcal{Y} = \{ y_t \}_{t=1}^{N}$. Following the noise model proposed in~\cite{brooks2019unprocessing,guo2021joint}, the noisy frame $y_t$ can be modeled as:
\begin{equation}
    y_t = M(x_t)+n(M(x_t),\sigma_s,\sigma_r),
\label{eq:degrademodel}
\end{equation}
where $n(x, \sigma_s, \sigma_r) \sim \mathcal{N}(0, \sigma_s x + \sigma_r^2)$ is a Gaussian noise model with shot noise scale $\sigma_s$ and read noise scale $\sigma_r$, and $M(\cdot)$ denotes the mosaic downsampling operator. Compared to the JDD for a single image or burst images, VJDD also needs to keep the temporal consistency of video frames. To address this challenge, we propose a consistent latent space propagation framework with temporal regularizations for VJDD, as illustrated in \cref{fig:Overall} (a). 

Firstly, we extract shallow features $\{ I_t \}_{t=1}^{N}$ from the noisy frames $\{ y_t \}_{t=1}^{N}$ and their corresponding noise maps $\{ m_t \}_{t=1}^{N}$, which are formed by the standard deviations of signal-dependent Gaussian noise at each pixel. The extracted features are propagated by using our consistent latent space propagation strategy. Specifically, the features are aligned by the proposed global-to-local alignment module (GLAM), and then aggregated and fed into the reconstruction module (RCM), resulting in the reconstructed features $H_{t}$. The clean full color frames $\hat{x}_t$ are obtained by applying a $3\times 3$ convolution layer on $H_{t}$. The detailed structures of GLAM are shown in \cref{fig:Overall} (b), where we perform alignment in a global-to-local manner to increase the receptive filed of the alignment module, and the structure of RCM is shown in \cref{fig:Overall} (c), where we employ a 5-scale UNet with skip connections. The channel attention module (CAM) and spatial attention module (SAM) are used to enhance the feature extraction capability of the RCM, whose implementation details can be found in \cite{guo2021joint}. To ensure the temporal consistency, we introduce two loss functions, \ie, the data temporal consistency loss and the relational perceptual consistency loss, to promote the temporal coherence in the reconstructed video frames.

\subsection{Consistent Latent Space Propagation}
Propagation plays a crucial role in leveraging the information within a video sequence for restoration tasks. There are three types of commonly used methods: sliding-window propagation~\cite{isobe2020video,wang2019edvr,guo2021joint,guo2022differentiable}, recurrent propagation~\cite{sajjadi2018frame,isobe2020video2}, bidirectional recurrent propagation~\cite{chan2020basicvsr,huang2015bidirectional,huang2017video,chan2022basicvsr++}. Sliding-window propagation restricts the accessible information for reconstructing the $t$-th frame within a local neighborhood $(t-\lfloor n/2 \rfloor:t+ \lfloor n/2 \rfloor)$, where $n$ denotes the number of frames. Such an approach effectively utilizes the information of local frames but omits distant frames, and the reconstruction of frame $t$ is independent of that of frame $t-1$, which is prone to temporal inconsistency. Recurrent propagation sequentially aggregate information, and the reconstruction of frame $t$ depends on the reconstructed frame $t-1$, hence promoting temporal consistency. Nonetheless, this approach only exploits the information from previous frames and overlooks subsequent frames. Bidirectional recurrent propagation leverages information from both forward and backward frames; however, this strategy requires to store the hidden state of all frames, resulting in high memory usage for video processing. 

To efficiently utilize the redundant information among video frames and maintain temporal consistency, we present a consistent latent space propagation scheme, as shown in \cref{fig:Overall} (a). We employ the features of neighboring frames $I_{t-\lfloor n/2 \rfloor:t+ \lfloor n/2 \rfloor}$, referred to as the buffer frames, and recurrently employ the reconstruction feature of previous frame, denoted by $H_{t-1}$, to reconstruct the desired frame $x_t$. The features of input frames are first aligned with the feature of reference frame $I_t$ by using the GLAM module, and then concatenated as the input to the RCM module. By incorporating the reconstruction results of previous frames as prior information, we can enhance the inter-frame consistency in the final output video.

The efficiency of our proposed propagation scheme lies in  two aspects. First, to restore the current frame $t$, we only need to store the features of buffer frames and the reconstruction feature of previous frame (\ie, $H_{t-1}$), resulting in a memory-efficient framework compared to bidirectional recurrent propagation~\cite{chan2020basicvsr,chan2022basicvsr++}. Second, when restoring the next frame $t+1$, the buffer frame features can be efficiently updated by removing the feature of frame $t-\lfloor n/2 \rfloor$ and adding the feature of frame $t+\lfloor n/2 \rfloor+1$. This allows us to reuse the features of $n-1$ frames without re-calculation. Different from the propagation method employed in RLSP~\cite{fuoli2019efficient}, our approach focuses on propagating the locally neighboring frames in latent space and reusing their features, resulting in much less computational cost. In addition, we design an alignment module to compensate for the offset between frames, which is crucial to enhance the video restoration performance  ~\cite{wang2019edvr,guo2022differentiable}.

\subsection{Global-to-Local Alignment Module}
Alignment is a crucial component for VJDD. While previous studies~\cite{chan2022basicvsr++,chan2021understanding} have demonstrated the effectiveness of deformable alignment in feature space~\cite{wang2019edvr,guo2021joint}, it encounters difficulties in dealing with large shift due to the limited receptive field~\cite{guo2022differentiable}. To address this issue, we design a GLAM module by first handling the global motion between features and then employing deformable convolution (DConv) to achieve pixel-wise alignment.

The structure of GLAM is illustrated in  \cref{fig:Overall} (b). By taking the reference feature $I_t$ and the target feature $I_{t+1}$ as inputs, GLAM aims to spatially align $I_{t+1}$ with $I_t$. To enhance the alignment performance, we adopt a pyramid alignment strategy with three scales. Firstly, we estimate the global motion at $\frac{1}{4}$ resolution to reduce the computational cost by solving the following equation:  
\begin{equation}
    \hat{f}_{g}^{\frac{1}{4}} = \operatorname*{argmin}_{f \in [-f_{max}, f_{max}]} \Vert \tau(I_{t+1}^{\frac{1}{4}}, f) - I_{t}^{\frac{1}{4}} \Vert,
\end{equation}
where $\tau(I, f)$ shifts image $I$ with the estimated global translation $f$, and $f_{max}$ is the maximum motion field. The estimated global motion is employed to align features at all scales by rescaling $\hat{f}_{g}^{\frac{1}{4}}$ to the feature resolution of other scales. Then, pixel alignment is performed using DConv. On each pyramidal scale $s$, suppose we have the globally aligned features $\hat{I}_{t}^s$ and $\hat{I}_{t+1}^s$. The offset is estimated as:
\begin{equation}
\hat{f}_{p}^s = g_1(\text{LReLU}([g_2([\hat{I}_{t}^s,\hat{I}_{t+1}^s]), \hat{f}_{p}^{(s-1) \uparrow2}])),
\end{equation}
where $g_1$ and $g_2$ are two convolutional layers, $[\cdot,\cdot]$ is the concatenation operator, and $(\cdot)^{\uparrow 2}$ is the upsampling operator with factor 2. The aligned feature ${I'}_{t+1}^s$ is obtained by:
\begin{equation}
    {I'}_{t+1}^s = g([\text{DConv}(\hat{I}_{t+1}^s, \hat{f}_{p}^s), {I'}_{t+1}^{(s-1) \uparrow2}]),
\end{equation}
where $g$ refers to general Conv+LReLU layer.

\subsection{Temporal Consistency Loss}
The flow-based temporal loss has been widely used in video processing to improve the temporal consistency ~\cite{lai2018learning,lei2019fully,lai2018learning,wang2020consistent,park2019preserving}. While those losses can address the temporal consistency issue of low-frequency contents in tasks like video colorization and video style transfer, they are not well-suited for preserving the temporal consistency of high-frequency textures and structures in video restoration tasks. In addition, flow-based temporal losses often lead to the over-smoothing problem~\cite{dai2022video} due to the flow estimation errors. To solve these limitations, we design two new temporal losses for the VJDD task.

\textbf{Data Temporal Consistency (DTC) Loss.}
The goal of DTC loss is to utilize the ground-truth optical flow to more precisely model the temporal consistency. During the training process, for a clean frame $x$, we randomly generate a sequence of optical flow $\{ f_{t\rightarrow t+1} \}_{t=1}^{N}$ from it. Then we synthesize a noisy raw video with the generated ground-truth optical flow by using $y_{t} = \mathcal{D}(\tau(x, f_{0\rightarrow t+1}),\sigma_s,\sigma_c)$, where $\mathcal{D}$ represents the degradation process described in \cref{eq:degrademodel}, and $f_{0\rightarrow t+1} = \sum_{k=0}^{t} f_{k\rightarrow k+1}$. Then the DTC loss can be defined as follows:
\begin{align}
    \mathcal{L}_{DTC} = \sum_{t=1}^{N-1}
    \Vert \tau(F(y_t), f_{t\rightarrow t+1}) - F(y_{t+1})\Vert,
\end{align}
where $F$ refers to our VJDD method. To account for long-term temporal consistency, we further introduce the long-term DTC loss, denoted as $\mathcal{L}_{DTC-l}$, as follows:
\begin{align}
    \mathcal{L}_{DTC-l} = \sum_{t=1}^{N-n}
    \Vert \tau(F(y_t), f_{t\rightarrow t+n}) - F(y_{t+n})\Vert.
\end{align}
By using the synthesized video to compute the temporal loss, $\mathcal{L}_{DTC}$ mitigates the error accumulation problem  of optical flow based temporal losses.

\textbf{Relational Perception Consistency (RPC) Loss.}
The DTC loss $\mathcal{L}_{DTC}$ is limited in capturing real motions because it utilizes simulated motions in training. We therefore further design a RPC loss, denoted by $\mathcal{L}_{RPC}$, which leverages real videos for training. The RPC loss aims to model the perceptual temporal relationship between the restoration results and the ground-truth video.  Like \cite{eilertsen2019single,dai2022video}, we directly apply the RPC loss on the unaligned frames. The $\mathcal{L}_{RPC}$ is defined as follows:
\begin{equation}
\small
    \mathcal{L}_{RPC} = \sum_{t=1}^{N-1}\Vert (\Phi(\hat{x}_t) - \Phi(\hat{x}_{t+1})) - (\Phi(x_t) - \Phi(x_{t+1}))\Vert,
\end{equation}
\noindent where $\hat{x}_t$ and $x_t$ represent the VJDD output and the ground-truth, respectively. The function $\Phi(\cdot)$ extracts perceptual features to promote consistency in structure and texture regions between frames. We adopt the pre-trained VGG perceptual network \cite{johnson2016perceptual} as $\Phi(\cdot)$. The term $(\Phi(x_t) - \Phi(x_{t+1}))$ captures the ground-truth perceptual relationship between clean frames, encompassing the unaligned intensity differences of the video. By comparing the perceptual relationship of ground-truth video and estimated video, the value of $\mathcal{L}_{RPC}$ can measure the presence of temporal flickering artifacts between frames.

{
\renewcommand{\arraystretch}{1.25}
\begin{table*}[!tbp]
\small
\centering
\caption{Quantitative comparison of different VJDD approaches on the REDS4 and 4KPix50 dataset. Following the experiment setting of \cite{mildenhall2018burst,xu2019learning}, ``Low" and ``High" noise levels are corresponding to $\sigma_s = 2.5\times 10^{-3}$, $\sigma_r = 10^{-2}$ and $\sigma_s = 6.4\times 10^{-3}$, $\sigma_r = 2\times 10^{-2}$, respectively. The best results are highlighted in bold.} 
\label{synthREDS4}
\begin{tabular}{c | c | c | c c c c c c c}
\hline
Noise &Dataset &Method &PSNR $\uparrow$ &SSIM $\uparrow$ &LPIPS $\downarrow$ &DISTS $\downarrow$ &WE ($10^{-2}$) $\downarrow$ &tOF $\downarrow$ &RWE ($10^{-2}$) $\downarrow$ \\
\hline
\multirow{10}{*}{Low} &\multirow{5}{*}{REDS4}
&EDVR-VJDD &34.02 &0.9105 &0.0964 &0.0670 &1.73 &0.3718 &1.42  \\
& &RviDeNet-VJDD &34.86 &0.9221 &0.0884 &0.0636 &1.78 &0.3742 &1.46  \\
& &GCP-Net &36.20 &0.9383 &0.0845 &0.0584 &1.30 &0.2889 &1.08  \\
& &2StageAlign &36.59 &\textbf{0.9448} &0.0791 &0.0553 &1.29 &0.2727 &1.07 \\
\cline{3-10}
& &Ours &\textbf{36.78} &0.9445 &\textbf{0.0654} &\textbf{0.0317} &\textbf{1.21} &\textbf{0.2477} &\textbf{1.02}  \\

\cline{2-10}

&\multirow{5}{*}{4KPix50}
& EDVR-VJDD &36.36 &0.9415 &0.1006 &0.0776 &1.14 &0.4941 &0.99  \\
& &RviDeNet-VJDD &36.52 &0.9392 &0.0990 &0.0736 &1.20 &0.4626 &1.03  \\
& &GCPNet &38.13 &0.9503 &0.0859 &0.0631 &0.82 &\textbf{0.4166} &0.76  \\
& &2StageAlign &38.16 &0.9570 &0.0853 &0.0644 &0.84 &0.4279 &0.78  \\
\cline{3-10}
& &Ours &\textbf{38.66} &\textbf{0.9576} &\textbf{0.0739} &\textbf{0.0424} &\textbf{0.79} &0.4201 &\textbf{0.72}  \\

\hline
\multirow{10}{*}{High} &\multirow{5}{*}{REDS4}
&EDVR-VJDD &32.06 &0.8715 &0.1421 &0.0887 &1.94 &0.5433 &1.74  \\
& &RviDeNet-VJDD &32.62 &0.8850 &0.1376 &0.0890 &2.02 &0.5681 &1.81  \\
& &GCP-Net &34.08 &0.9124 &0.1297 &0.0817 &1.36 &0.4442 &1.28 \\
& &2StageAlign &34.43 &0.9178 &0.1257 &0.0757 &1.39 &0.4476 &1.30  \\
\cline{3-10}
& &Ours &\textbf{34.72} &\textbf{0.9181} &\textbf{0.0977} &\textbf{0.0474} &\textbf{1.23} &\textbf{0.3692} &\textbf{1.17}  \\

\cline{2-10}

&\multirow{5}{*}{4KPix50}
&EDVR-VJDD &34.62 &0.9188 &0.1415 &0.1033 &1.28 &0.6349 &1.20  \\
& &RviDeNet-VJDD &34.67 &0.9147 &0.1451 &0.1082 &1.38 &0.6203 &1.28  \\
& &GCP-Net &36.13 &0.9411 &0.1132 &0.0815 &0.88 &0.5594 &0.90  \\
& &2StageAlign &36.27 &0.9424 &0.1114 &0.0792 &0.89 &0.5600 &0.88  \\
\cline{3-10}
& &Ours &\textbf{36.96} &\textbf{0.9430} &\textbf{0.1002} &\textbf{0.0569} &\textbf{0.80} &\textbf{0.5399} &\textbf{0.82}  \\

\hline
\end{tabular} 
\label{table:reds4}
\end{table*}
}

\subsection{Overall Training Loss}
Following previous JDD methods~\cite{guo2021joint,guo2022differentiable}, we first define a reconstruction loss $\mathcal{L}_{r}$ in both linear RGB domain and sRGB domain:
\begin{equation}
\small
    \mathcal{L}_{r} = \sum_{t=1}^N \sqrt{\Vert \hat{x}_t - x_t \Vert^2 + \epsilon^2} + \sqrt{\Vert \Gamma(\hat{x}_t) - \Gamma(x_t) \Vert^2 + \epsilon^2},
\end{equation}
where $\Gamma(\cdot)$ is a predefined ISP operator and we adopt the one in \cite{brooks2019unprocessing}. Additionally, to obtain visually more pleasing VJDD results, the perceptual loss \cite{johnson2016perceptual} is also used:
\begin{equation}
    \mathcal{L}_{p} = \sum_{t=1}^N \sqrt{\Vert \Phi(\Gamma(\hat{x}_t)) - \Phi(\Gamma(x_t)) \Vert^2 + \epsilon^2}.
\end{equation}
The overall loss function is:
\begin{equation}
    \mathcal{L} = \mathcal{L}_{r} + \lambda\mathcal{L}_{p} + \alpha\mathcal{L}_{DTC} + \beta\mathcal{L}_{DTC-l} + \gamma\mathcal{L}_{RPC},
\end{equation}
where $\lambda$, $\alpha$, $\beta$ and $\gamma$ are  balance parameters, and we empirically set them as 0.002, 0.5, 0.2 and 0.001, respectively, in all our experiments.

\begin{figure}[!tbp]
\centering
\includegraphics[width=0.4\textwidth]{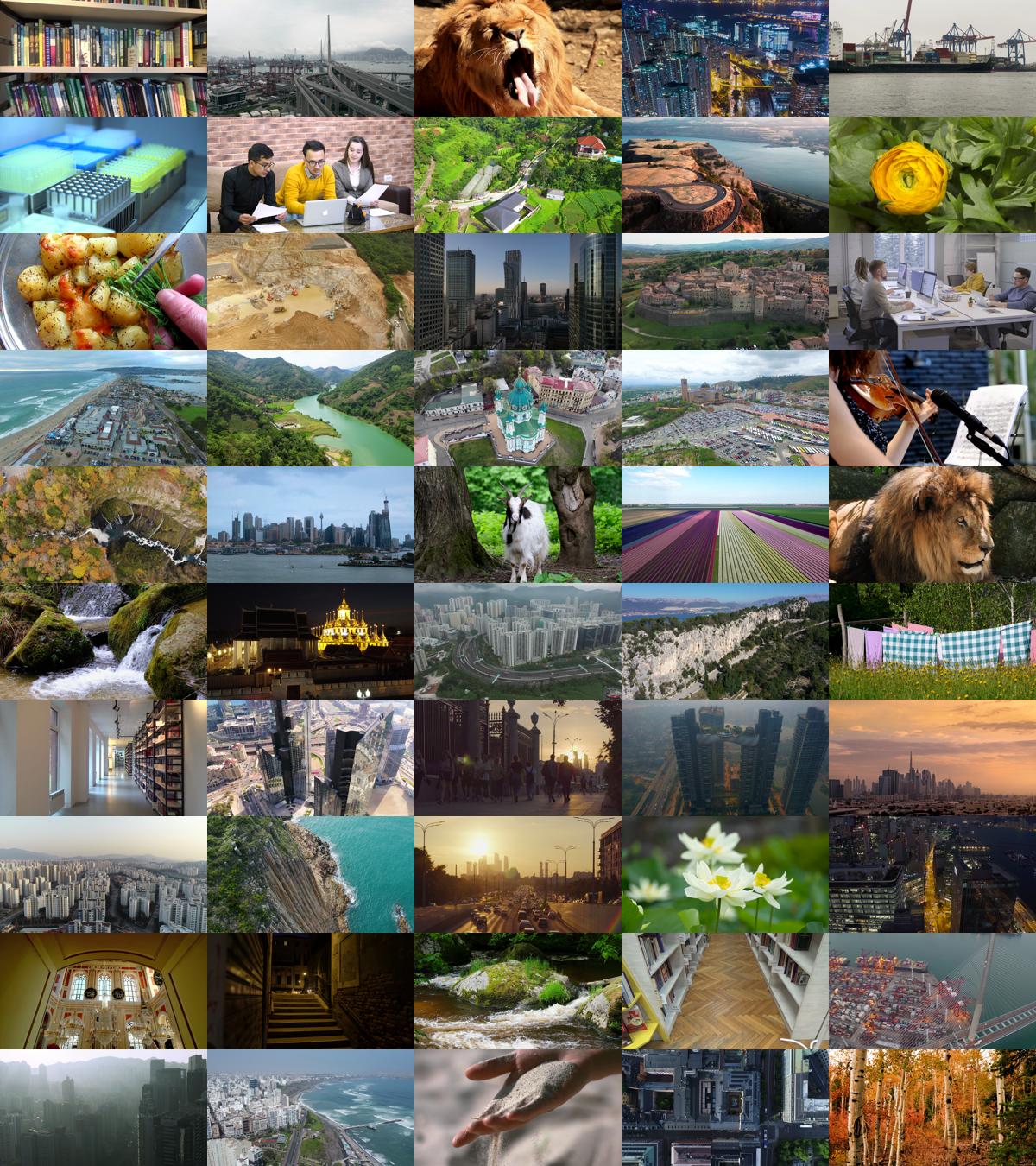}\vspace{0.2cm}

\begin{overpic}[width=0.35\textwidth]{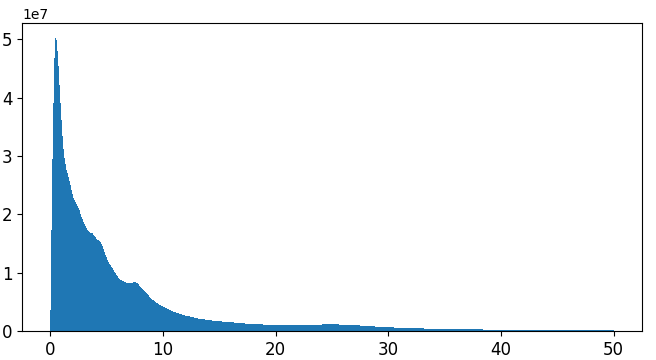}
    \put(-5.0,15){\footnotesize \rotatebox{90}{Frequency}}
    \put(48,-4){\footnotesize Shift}
\end{overpic}

\caption{Top: scenes of our collected 4KPix50 videos. Bottom: distribution of motions of the 4KPix50 dataset.}
\label{fig:all_4K}
\end{figure}

\begin{figure*}[!h]
\centering
\begin{minipage}[b]{1.0\textwidth}
\centering
    \begin{minipage}[b]{0.24\textwidth}
    \centering
    \includegraphics[width=1\textwidth]{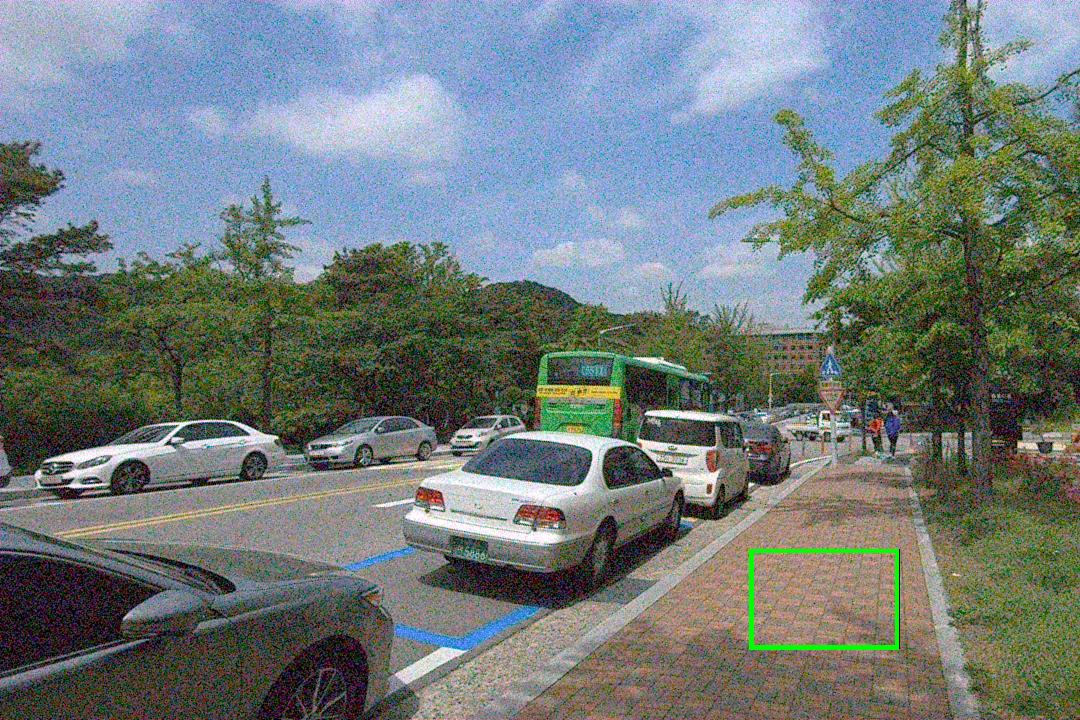}
    {\small Noise image (\emph{Clip000})}
    \end{minipage} 
    \begin{minipage}[b]{0.24\textwidth}
    \centering
    \includegraphics[width=1\textwidth]{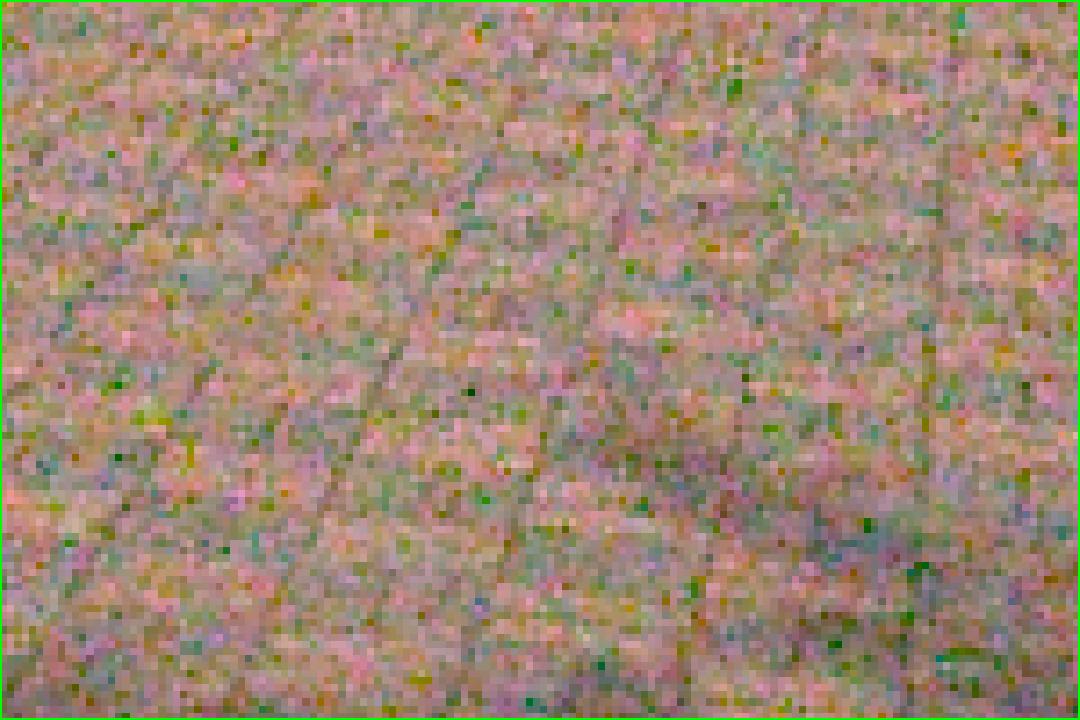}
    {\small Noise patch}
    \end{minipage} 
    \begin{minipage}[b]{0.24\textwidth}
    \centering
    \includegraphics[width=1\textwidth]{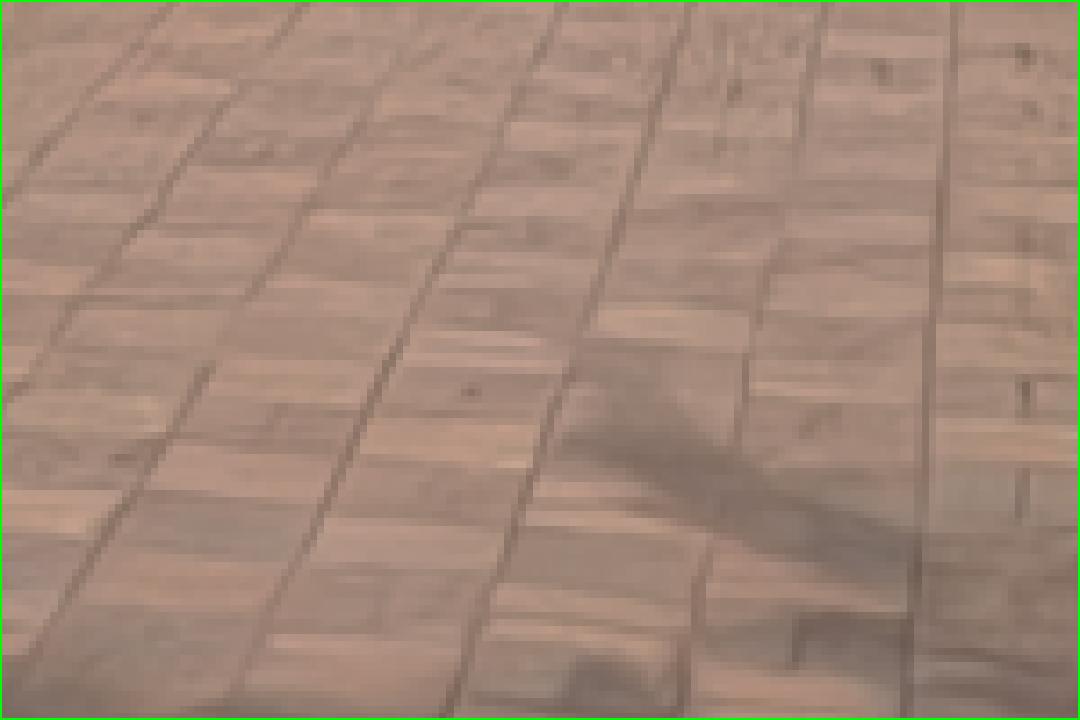}
    {\small EDVR-VJDD~\cite{wang2019edvr}}
    \end{minipage} 
    \begin{minipage}[b]{0.24\textwidth}
    \centering
    \includegraphics[width=1\textwidth]{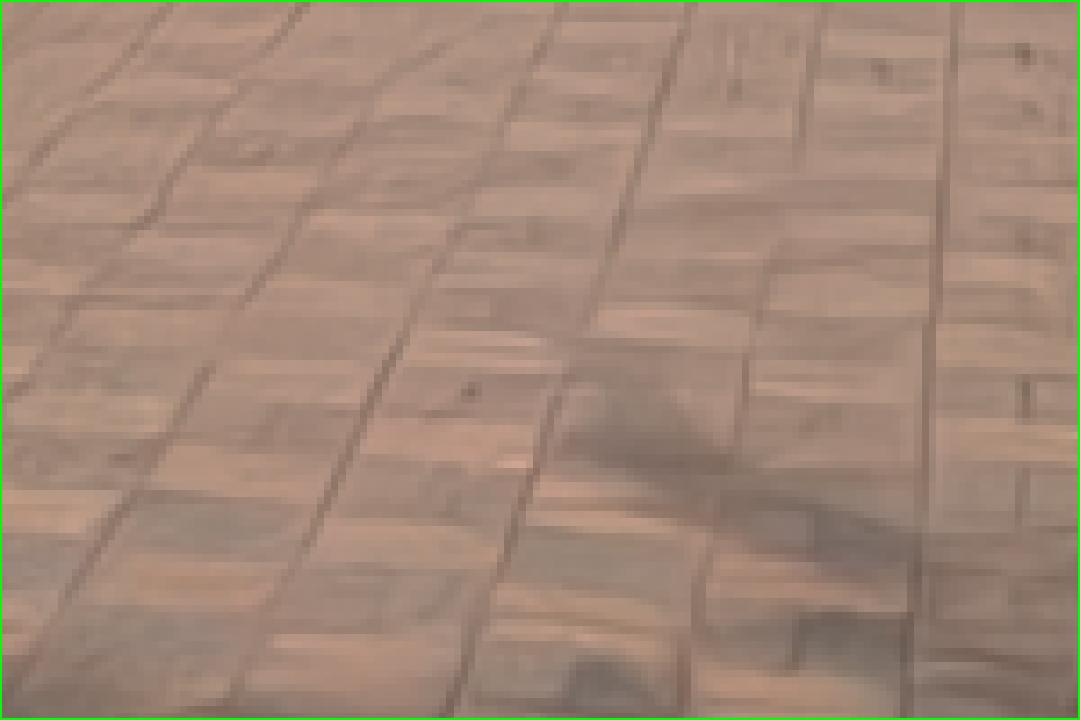}
    {\small RviDeNet-VJDD~\cite{yue2020supervised}}
    \end{minipage} 
\end{minipage}
\vspace{0.15cm}

\begin{minipage}[b]{1.0\textwidth}
\centering
    \begin{minipage}[b]{0.24\textwidth}
    \centering
    \includegraphics[width=1\textwidth]{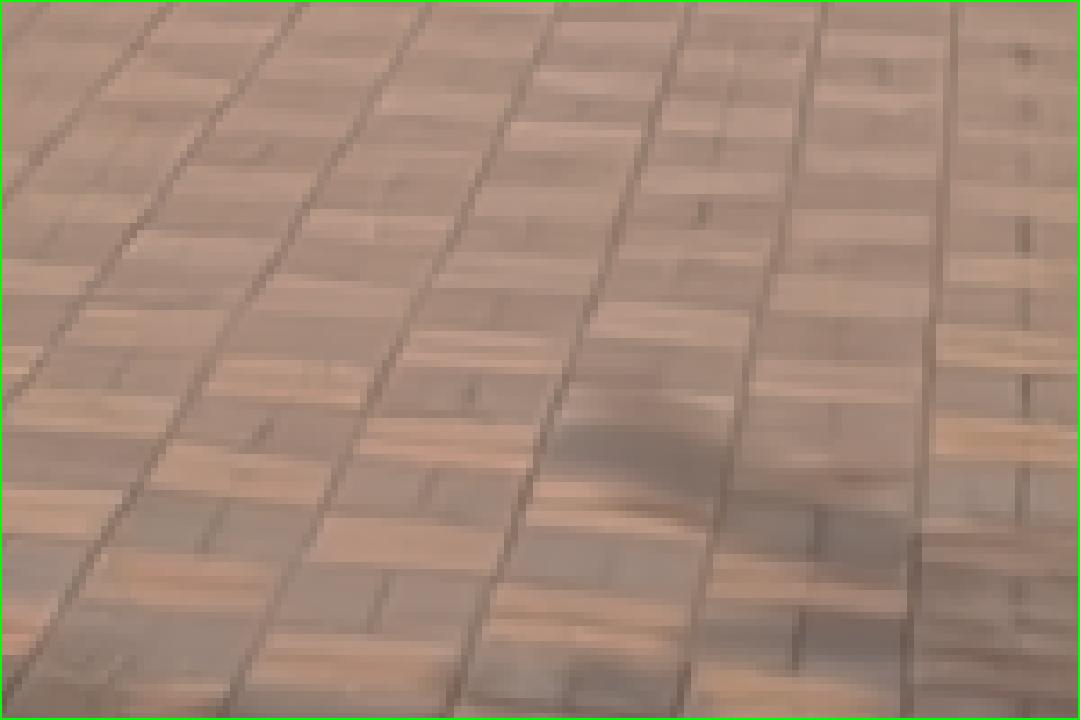}
    {\small GCP-Net~\cite{guo2021joint}}
    \end{minipage} 
    \begin{minipage}[b]{0.24\textwidth}
    \centering
    \includegraphics[width=1\textwidth]{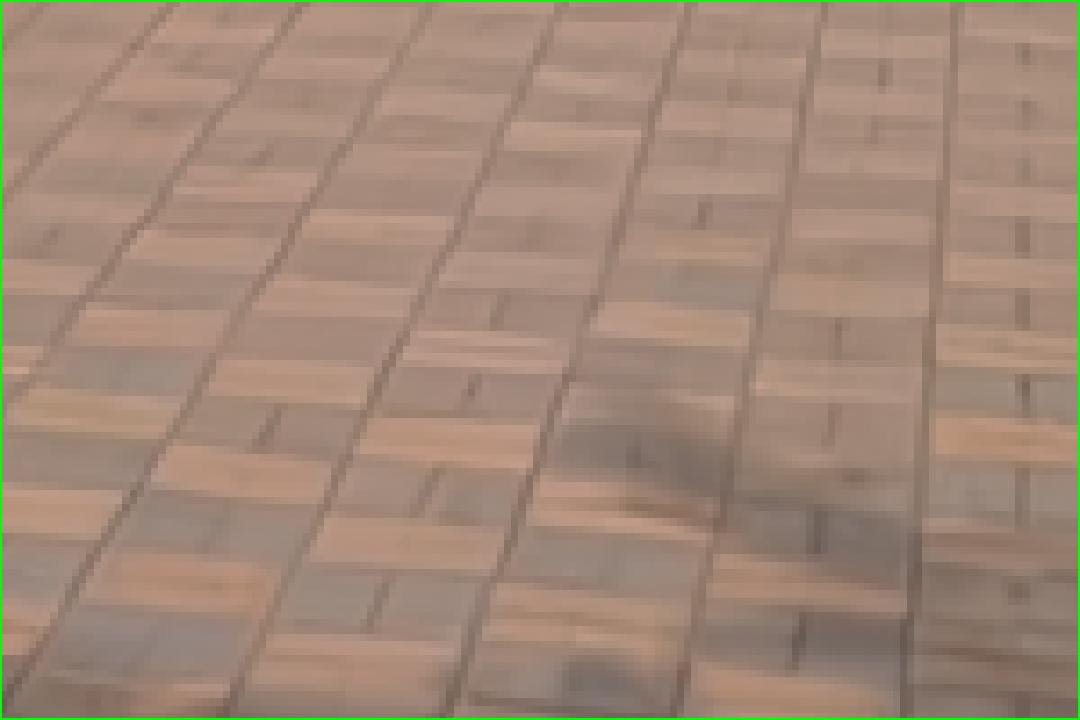}
    {\small 2StageAlign~\cite{guo2022differentiable}}
    \end{minipage} 
    \begin{minipage}[b]{0.24\textwidth}
    \centering
    \includegraphics[width=1\textwidth]{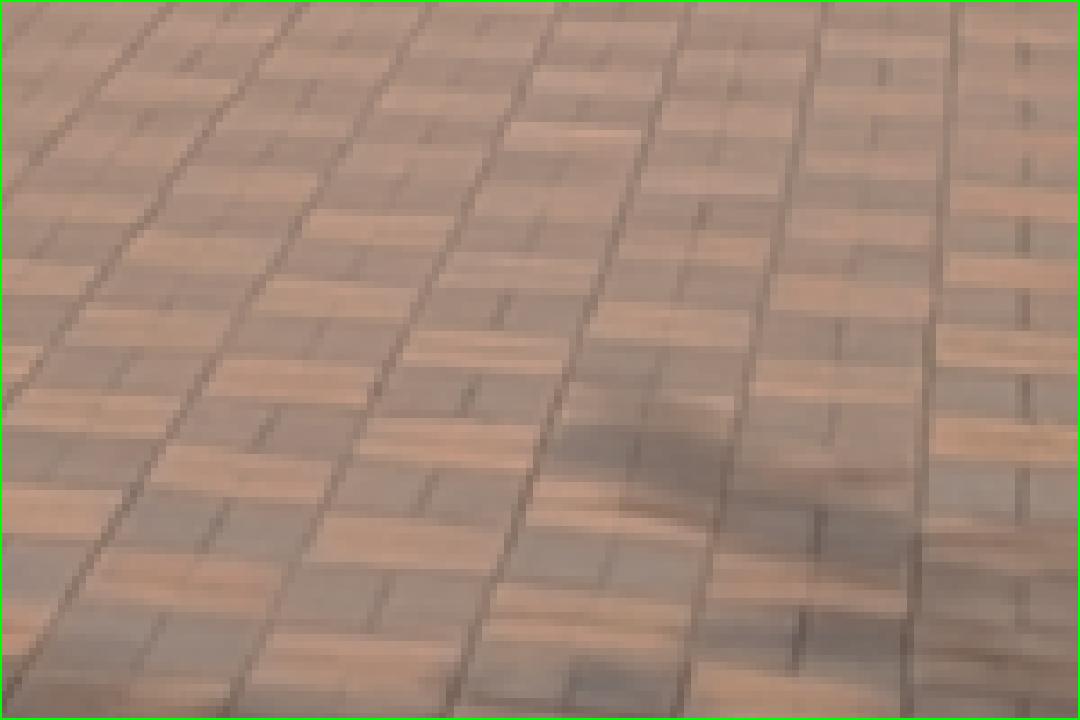}
    {\small (g) Ours}
    \end{minipage} 
    \begin{minipage}[b]{0.24\textwidth}
    \centering
    \includegraphics[width=1\textwidth]{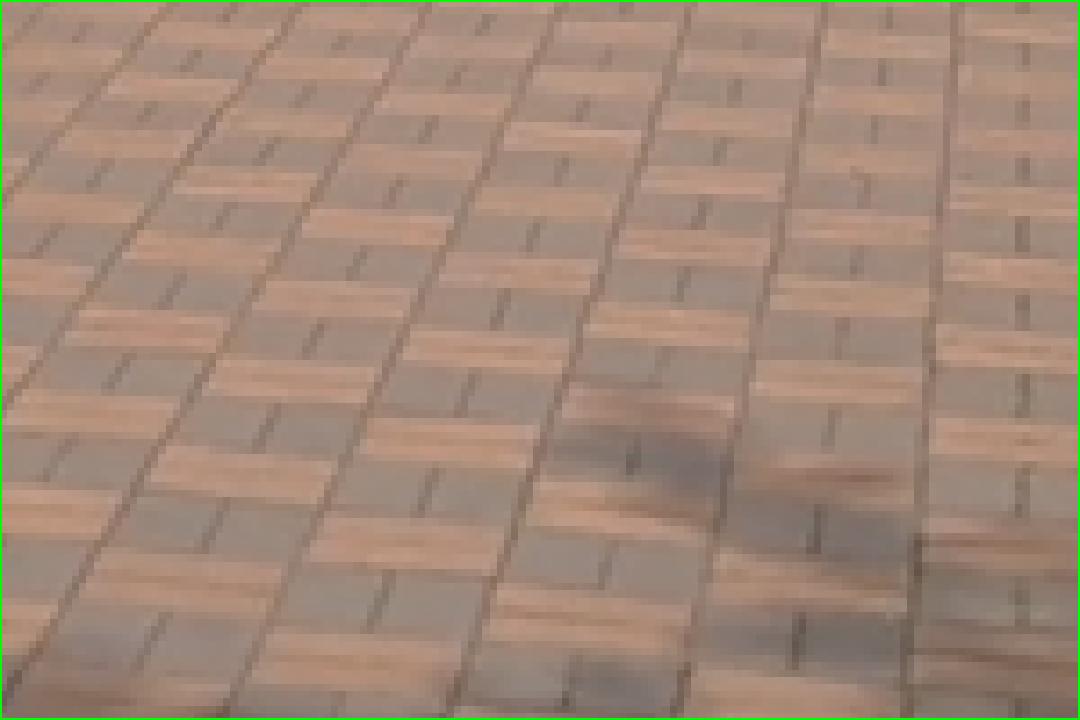}
    {\small (h) GT}
    \end{minipage} 
\end{minipage}\vspace{0.15cm}

\begin{minipage}[b]{1.0\textwidth}
\centering
    \begin{minipage}[b]{0.24\textwidth}
    \centering
    \includegraphics[width=1\textwidth]{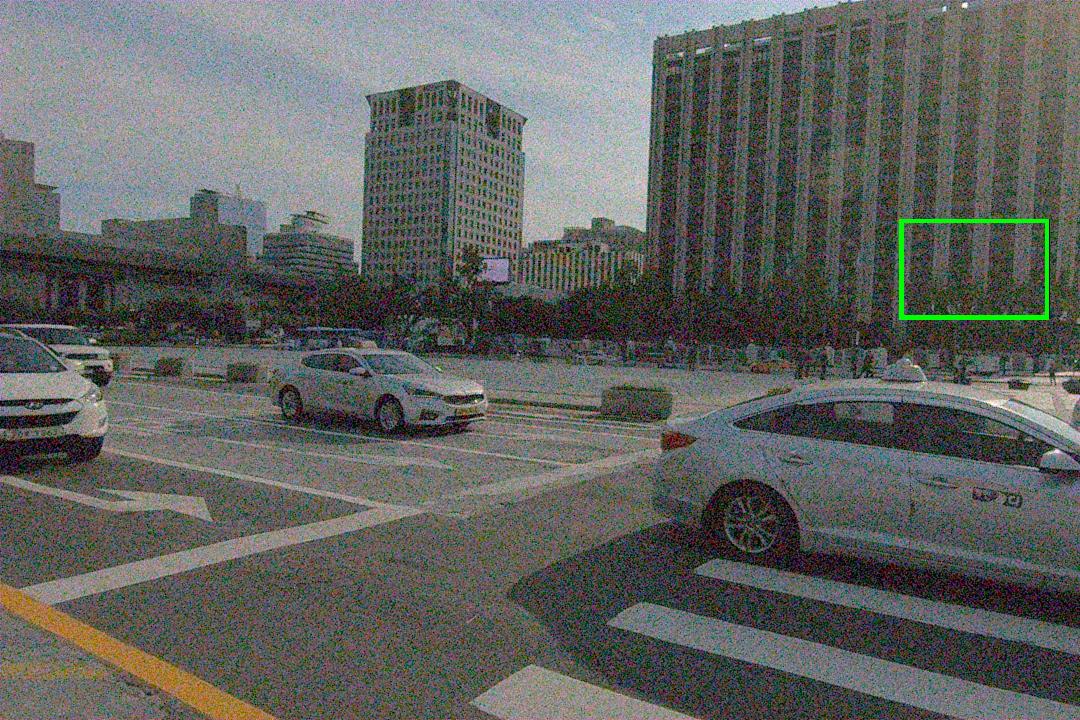}
    {\small Noise image (\emph{Clip015})}
    \end{minipage} 
    \begin{minipage}[b]{0.24\textwidth}
    \centering
    \includegraphics[width=1\textwidth]{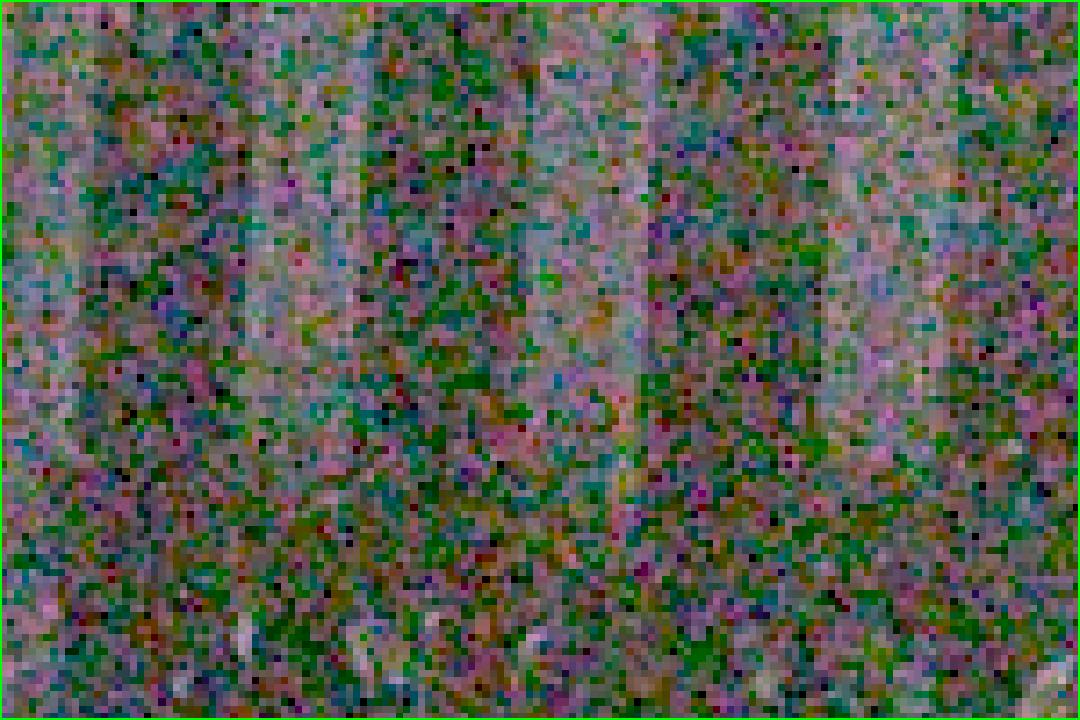}
    {\small Noise patch}
    \end{minipage} 
    \begin{minipage}[b]{0.24\textwidth}
    \centering
    \includegraphics[width=1\textwidth]{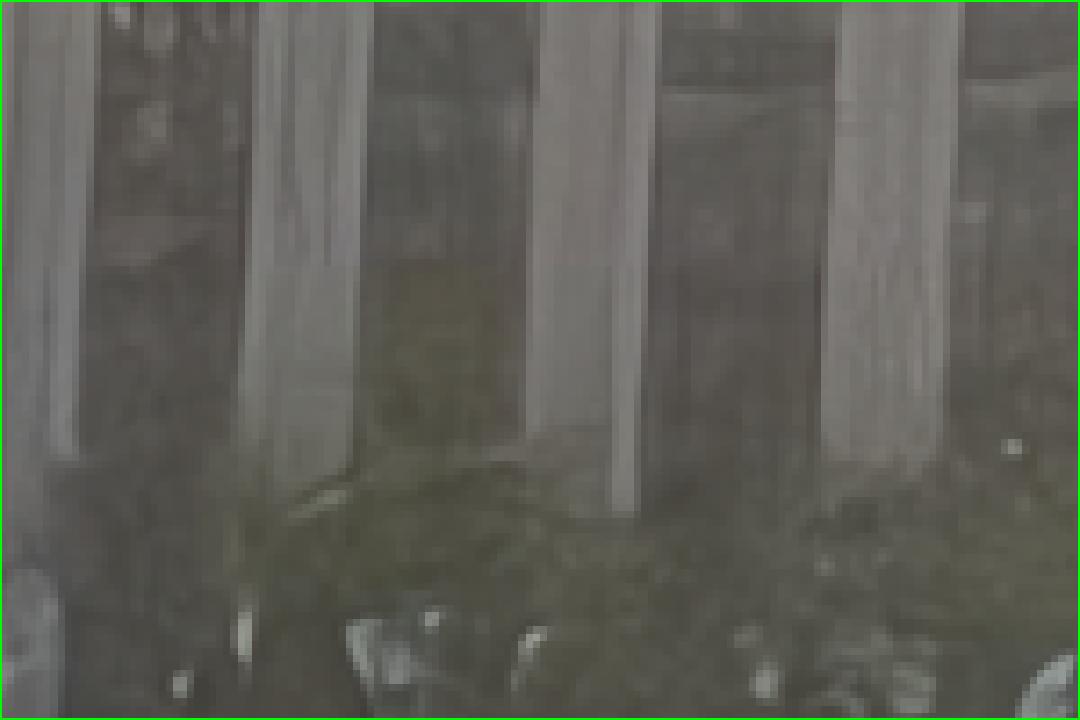}
    {\small EDVR-VJDD~\cite{wang2019edvr}}
    \end{minipage} 
    \begin{minipage}[b]{0.24\textwidth}
    \centering
    \includegraphics[width=1\textwidth]{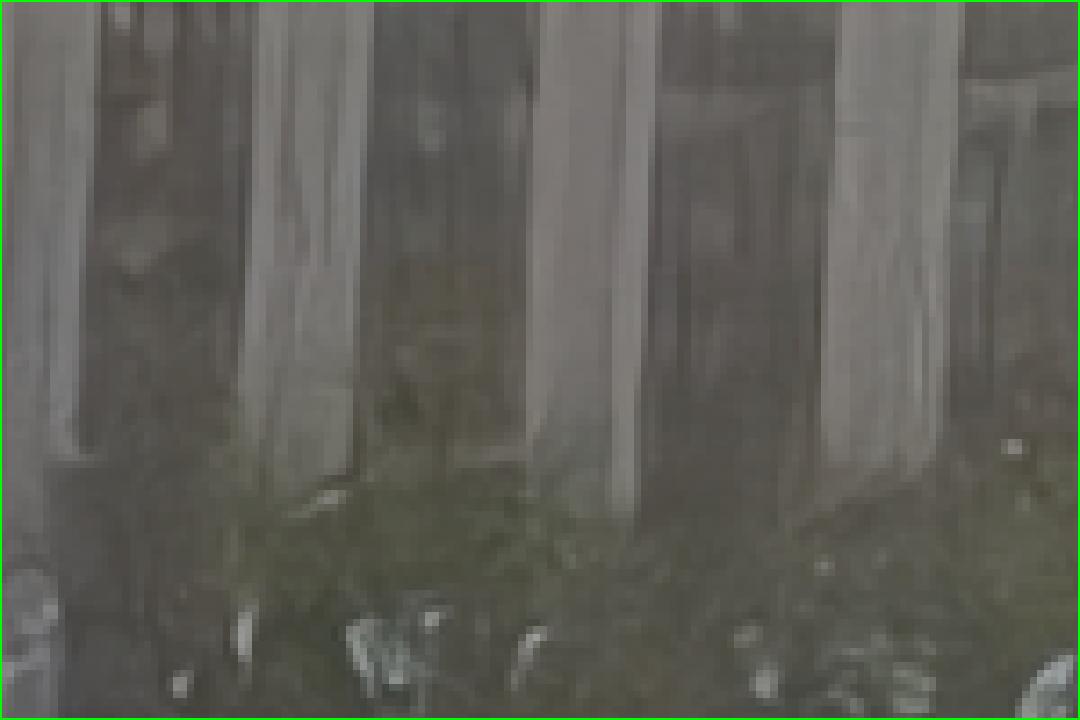}
    {\small RviDeNet-VJDD~\cite{yue2020supervised}}
    \end{minipage} 
\end{minipage}
\vspace{0.15cm}

\begin{minipage}[b]{1.0\textwidth}
\centering
    \begin{minipage}[b]{0.24\textwidth}
    \centering
    \includegraphics[width=1\textwidth]{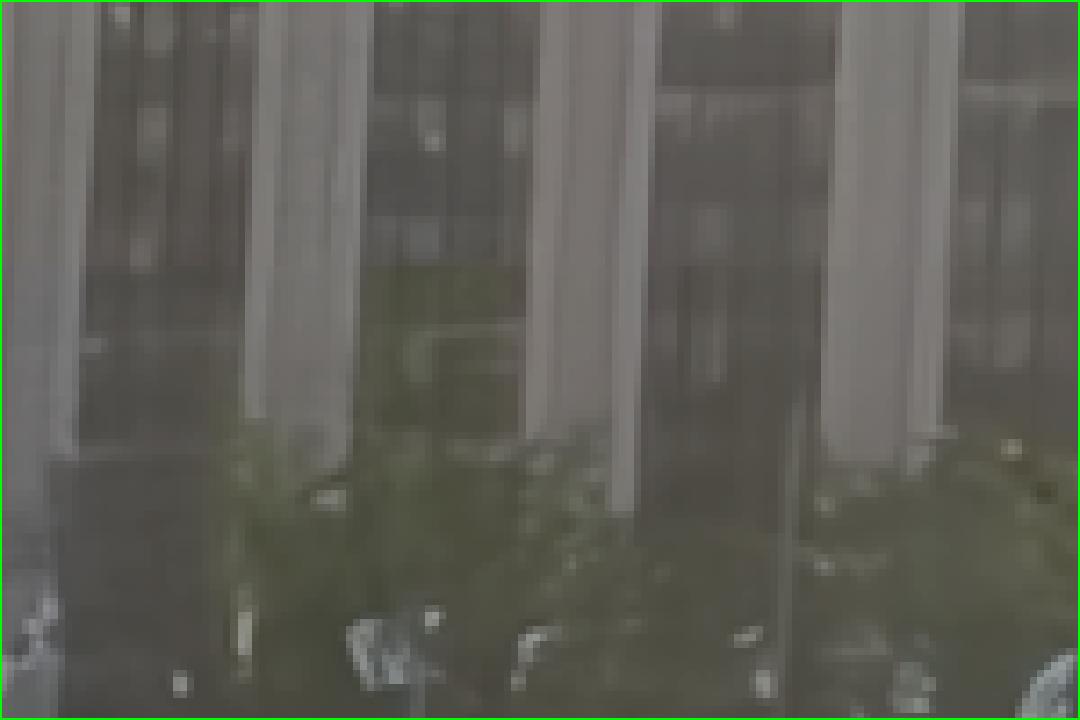}
    {\small GCP-Net~\cite{guo2021joint}}
    \end{minipage} 
    \begin{minipage}[b]{0.24\textwidth}
    \centering
    \includegraphics[width=1\textwidth]{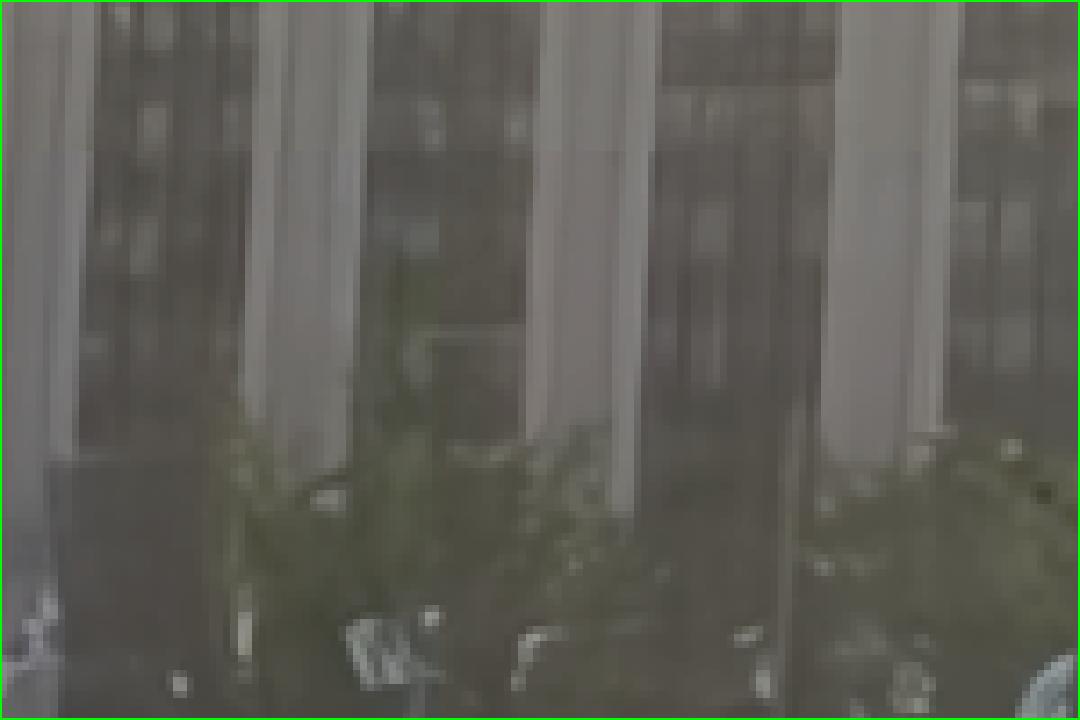}
    {\small 2StageAlign~\cite{guo2022differentiable}}
    \end{minipage} 
    \begin{minipage}[b]{0.24\textwidth}
    \centering
    \includegraphics[width=1\textwidth]{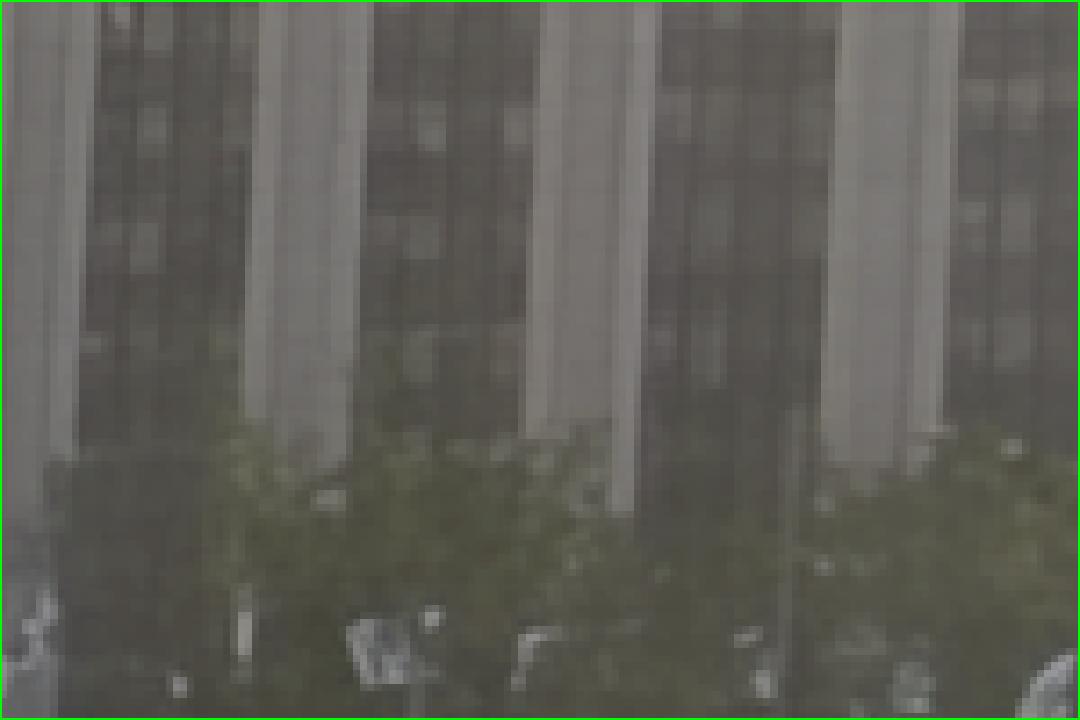}
    {\small Ours}
    \end{minipage} 
    \begin{minipage}[b]{0.24\textwidth}
    \centering
    \includegraphics[width=1\textwidth]{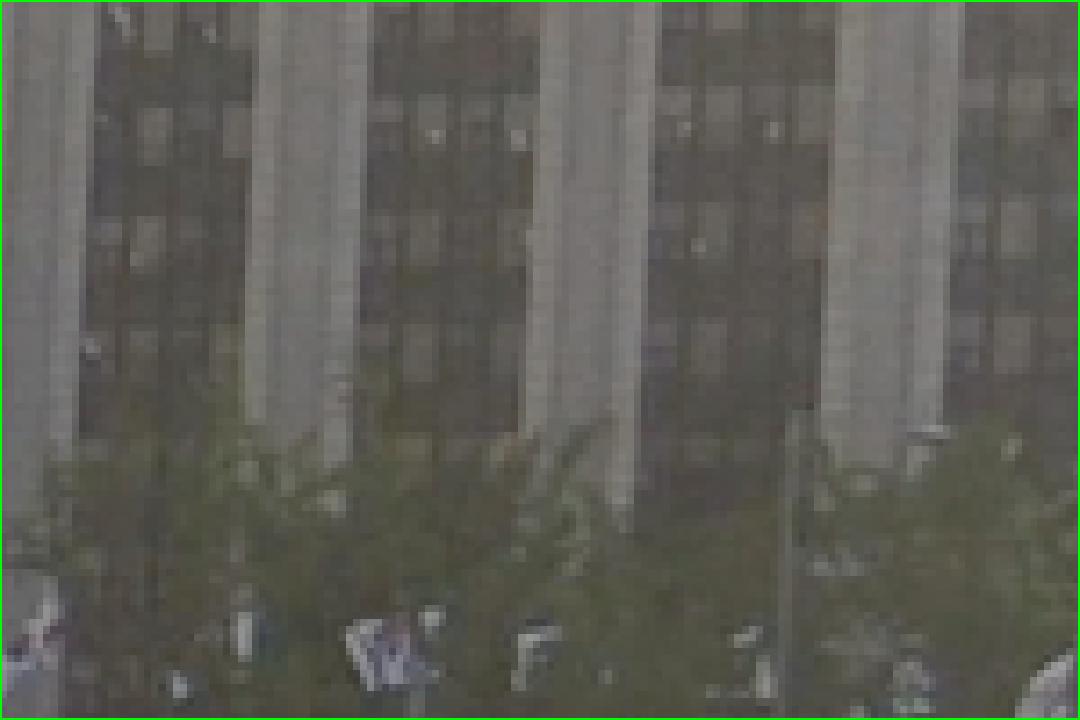}
    {\small GT}
    \end{minipage} 
\end{minipage}\vspace{0.15cm}

\begin{minipage}[b]{1.0\textwidth}
\centering
    \begin{minipage}[b]{0.24\textwidth}
    \centering
    \includegraphics[width=1\textwidth]{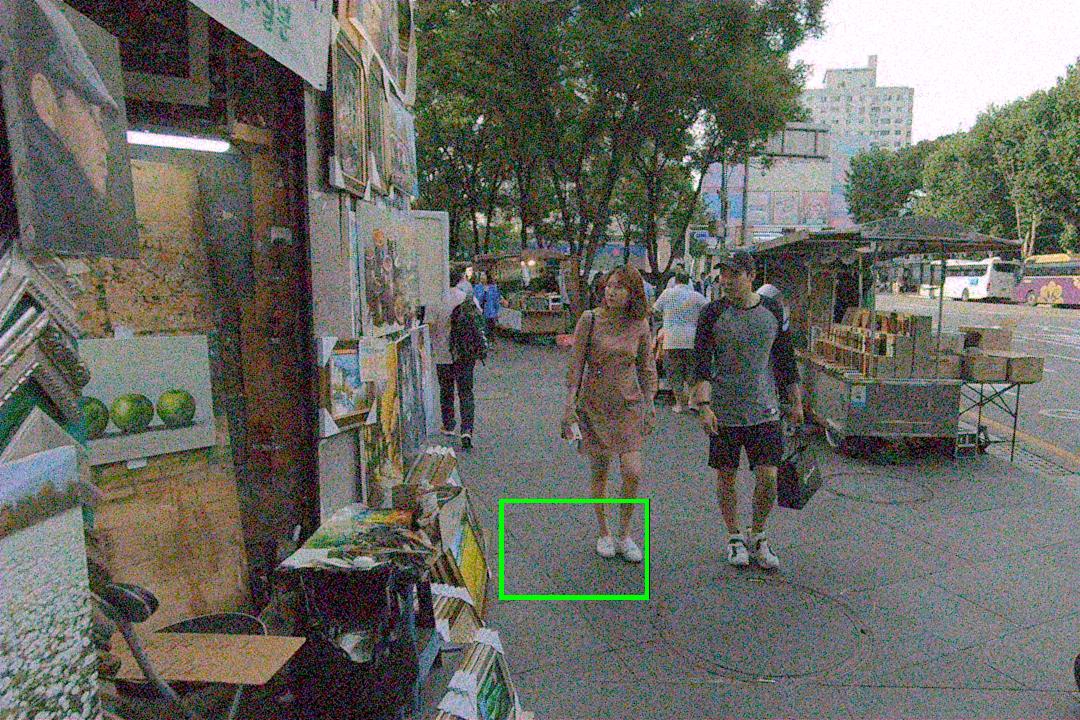}
    {\small Noisy image (\emph{Clip020})}
    \end{minipage} 
    \begin{minipage}[b]{0.24\textwidth}
    \centering
    \includegraphics[width=1\textwidth]{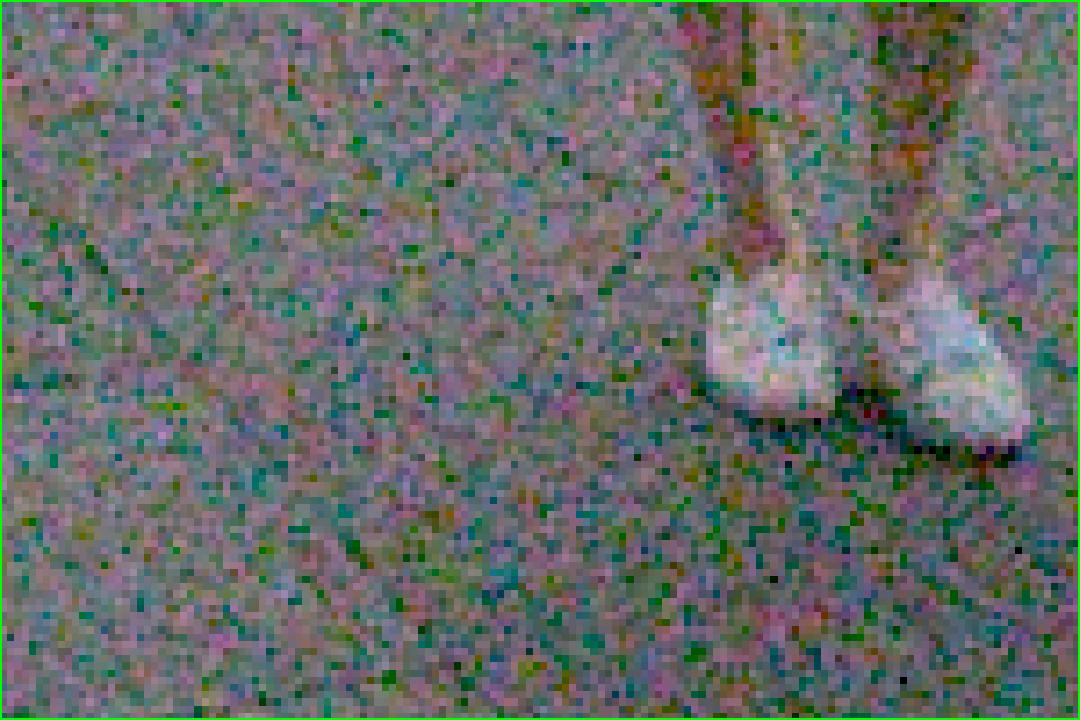}
    {\small Noisy patch}
    \end{minipage} 
    \begin{minipage}[b]{0.24\textwidth}
    \centering
    \includegraphics[width=1\textwidth]{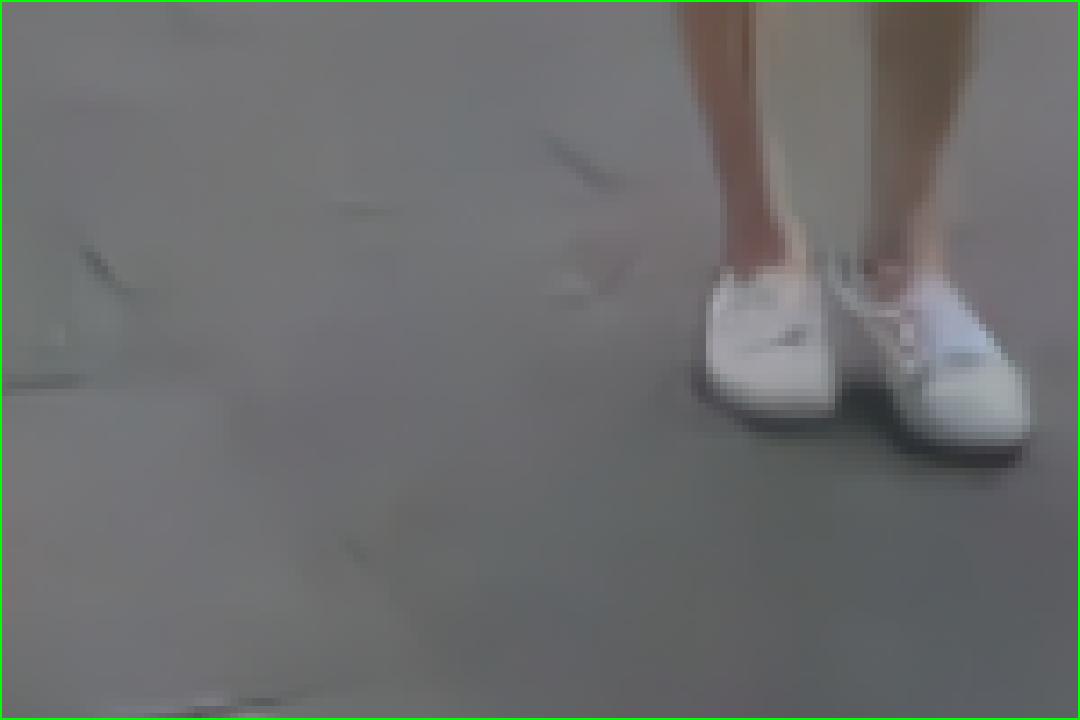}
    {\small EDVR-VJDD~\cite{wang2019edvr}}
    \end{minipage} 
    \begin{minipage}[b]{0.24\textwidth}
    \centering
    \includegraphics[width=1\textwidth]{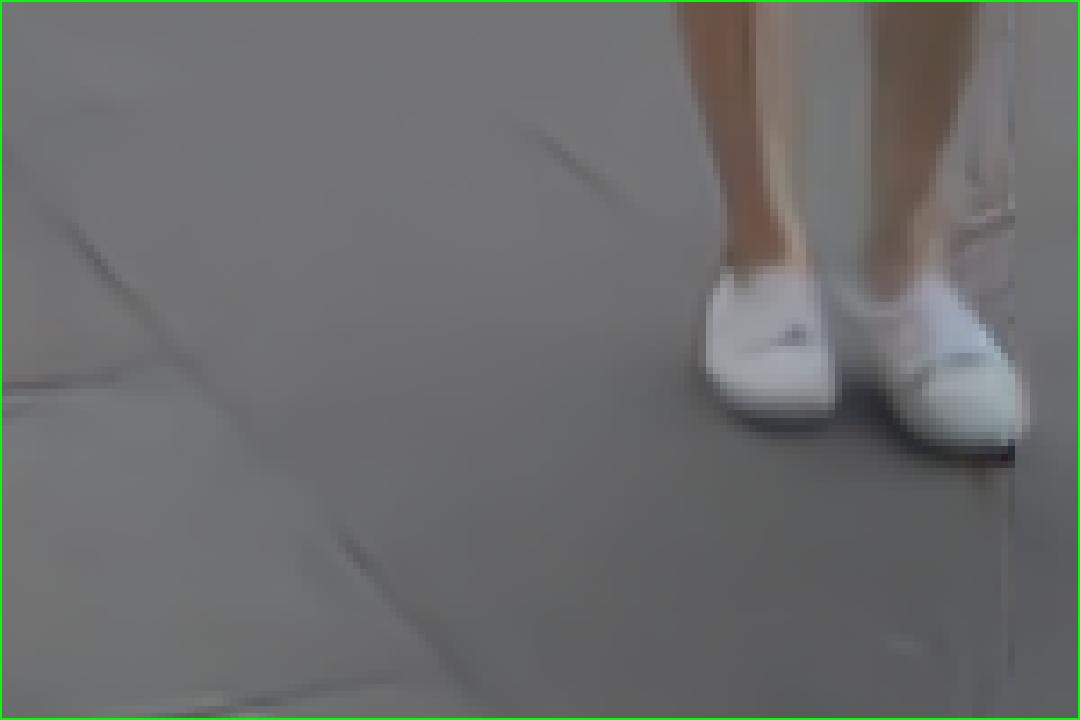}
    {\small RviDeNet-VJDD~\cite{yue2020supervised}}
    \end{minipage} 
\end{minipage}
\vspace{0.15cm}

\begin{minipage}[b]{1.0\textwidth}
\centering
    \begin{minipage}[b]{0.24\textwidth}
    \centering
    \includegraphics[width=1\textwidth]{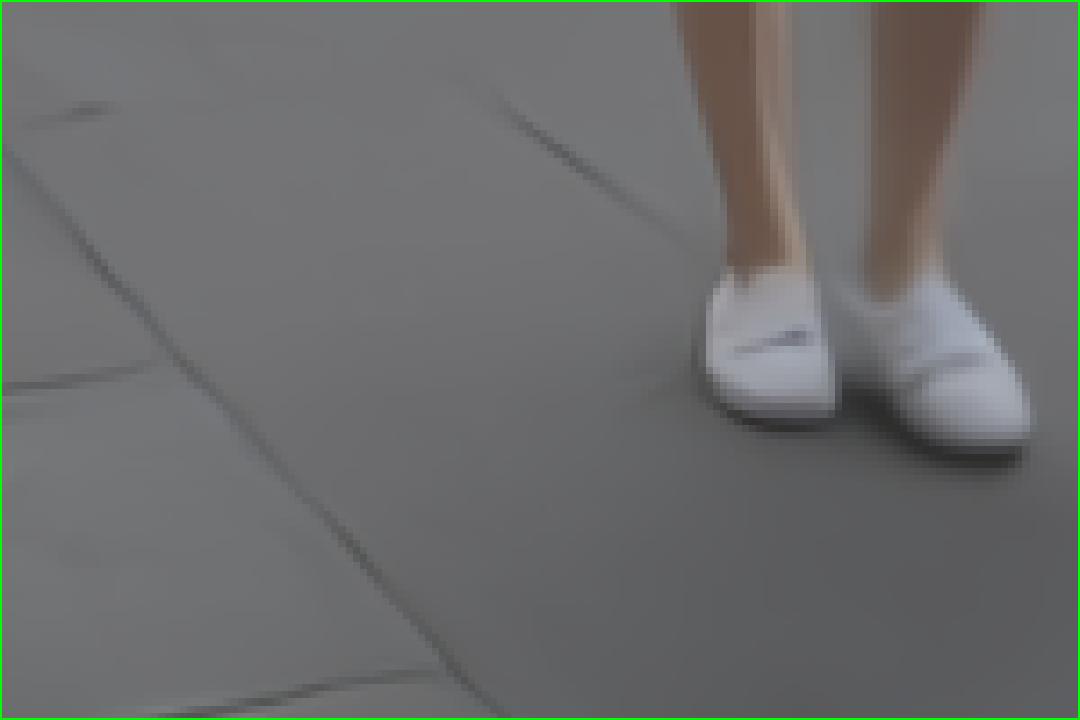}
    {\small GCP-Net~\cite{guo2021joint}}
    \end{minipage} 
    \begin{minipage}[b]{0.24\textwidth}
    \centering
    \includegraphics[width=1\textwidth]{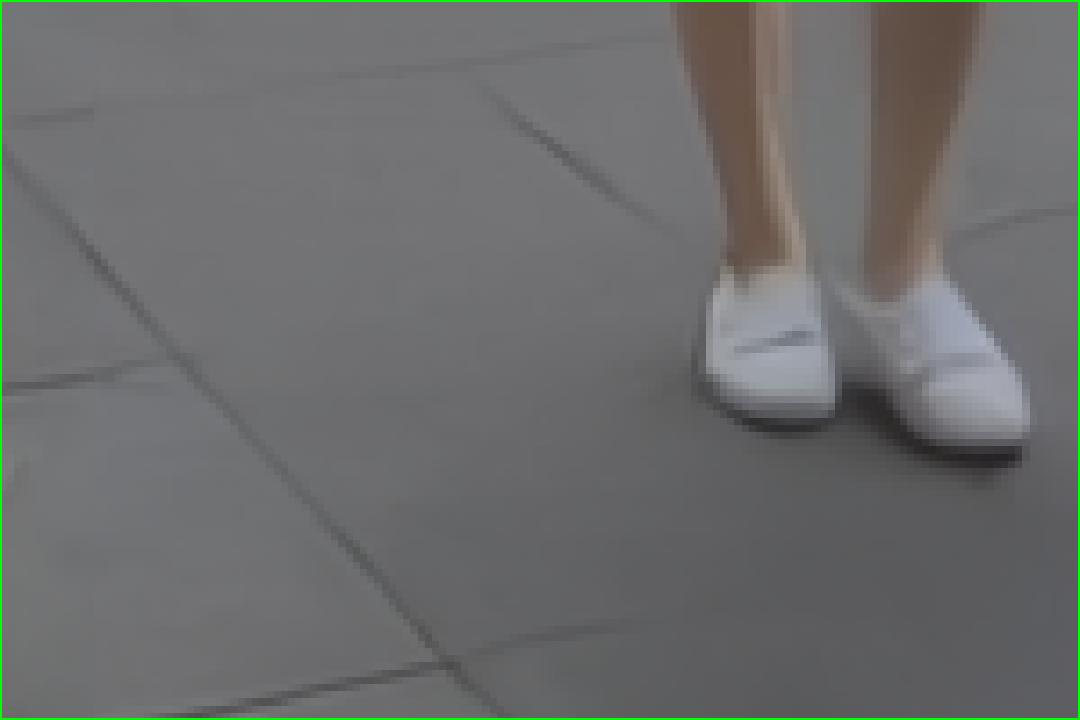}
    {\small 2StageAlign~\cite{guo2022differentiable}}
    \end{minipage} 
    \begin{minipage}[b]{0.24\textwidth}
    \centering
    \includegraphics[width=1\textwidth]{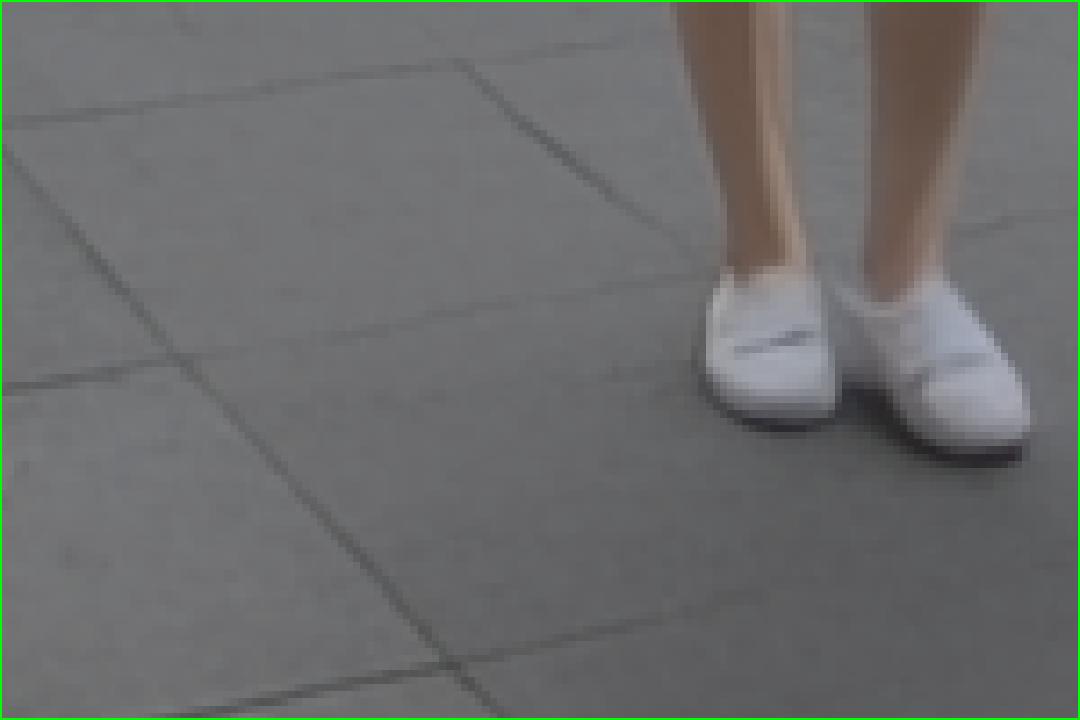}
    {\small Ours}
    \end{minipage} 
    \begin{minipage}[b]{0.24\textwidth}
    \centering
    \includegraphics[width=1\textwidth]{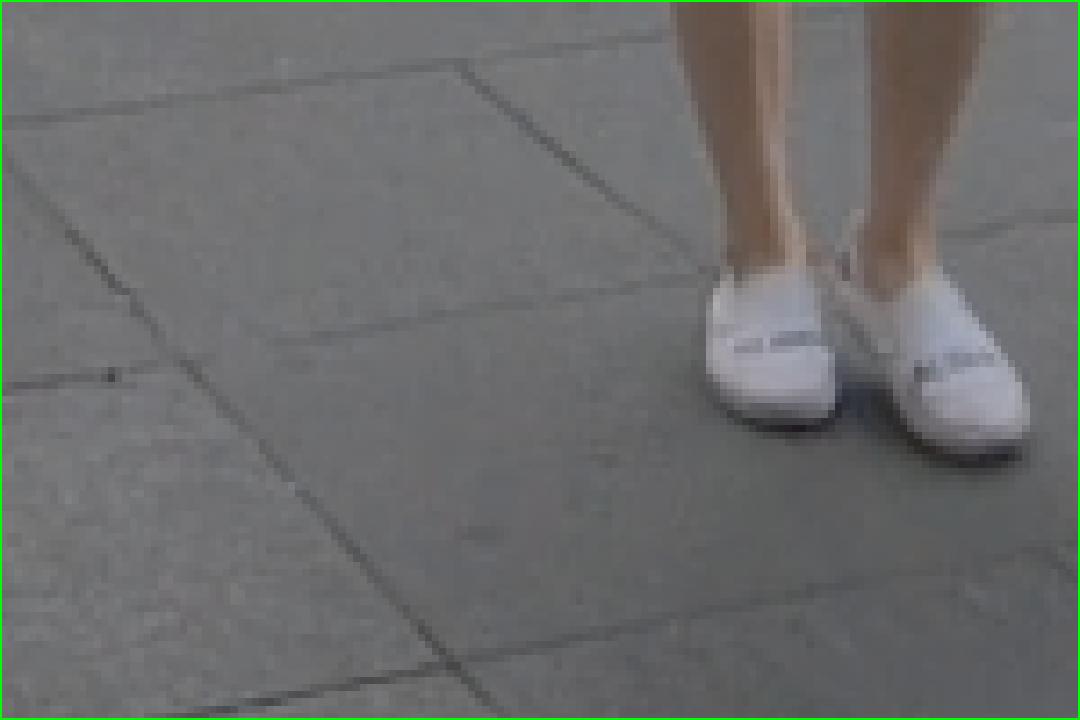}
    {\small GT}
    \end{minipage} 
\end{minipage}
\caption{VJDD results by different methods on the REDS4 dataset.}
\label{fig:Reds4}
\end{figure*}


\section{Experiments}
\subsection{Experiment Setting}
\textbf{Training Details.} We adhere to the experimental setup in \cite{guo2021joint,guo2022differentiable} and utilize the REDS dataset~\cite{nah2019ntire} for training. The REDS dataset comprises 240 training clips with a resolution of 720p. We use \cref{eq:degrademodel} to generate noisy raw videos, where the noise parameters $\sigma_s$ and $\sigma_r$ are uniformly sampled from the ranges of $[10^{-4}, 10^{-2}]$ and $[10^{-3}, 10^{-1.5}]$, respectively. Considering that the quality of estimated videos at the early stages is poor, applying the temporal losses $\mathcal{L}_{DTC}$ and $\mathcal{L}_{RPC}$ at the beginning of training is not useful. Therefore, we adopt a two-step training strategy. First, we obtain a pre-trained model using only the reconstruction loss $\mathcal{L}_{r}$. Then, we train the model by using the full loss  $\mathcal{L}$ based on the pre-trained model. Our model is implemented using PyTorch and trained on two RTX 3090Ti GPUs. The Adam optimizer is employed with a momentum of 0.9~\cite{kingma2014adam}. The learning rate is initialized as $1\times 10^{-4}$ and is decreased using the cosine function~\cite{loshchilov2016sgdr}.

\begin{figure*}[!h]
\centering
\begin{minipage}[b]{1.0\textwidth}
\centering
    \begin{minipage}[b]{0.16\textwidth}
    \centering
    \includegraphics[width=1\textwidth]{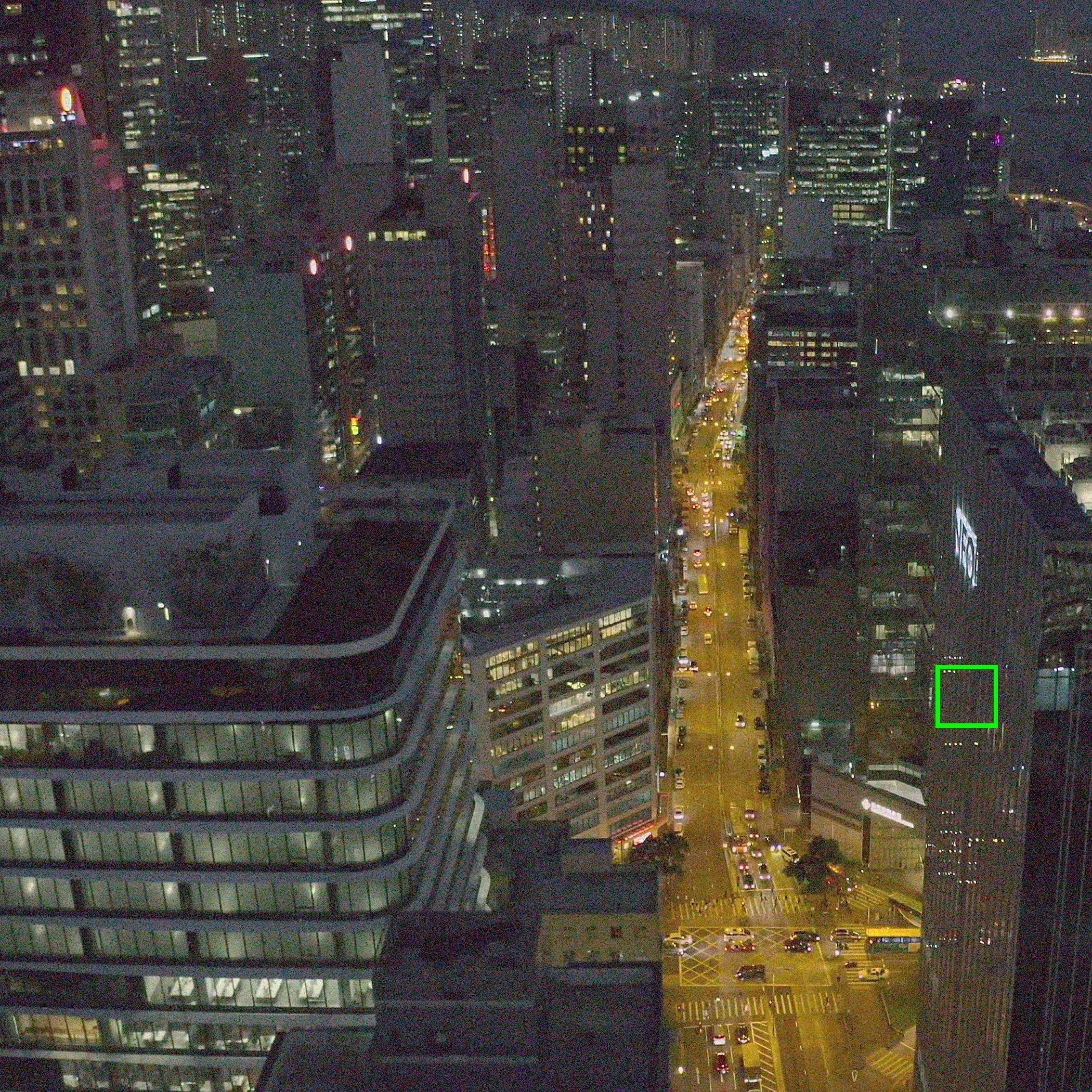}
    \end{minipage} 
    \begin{minipage}[b]{0.16\textwidth}
    \centering
    \includegraphics[width=1\textwidth]{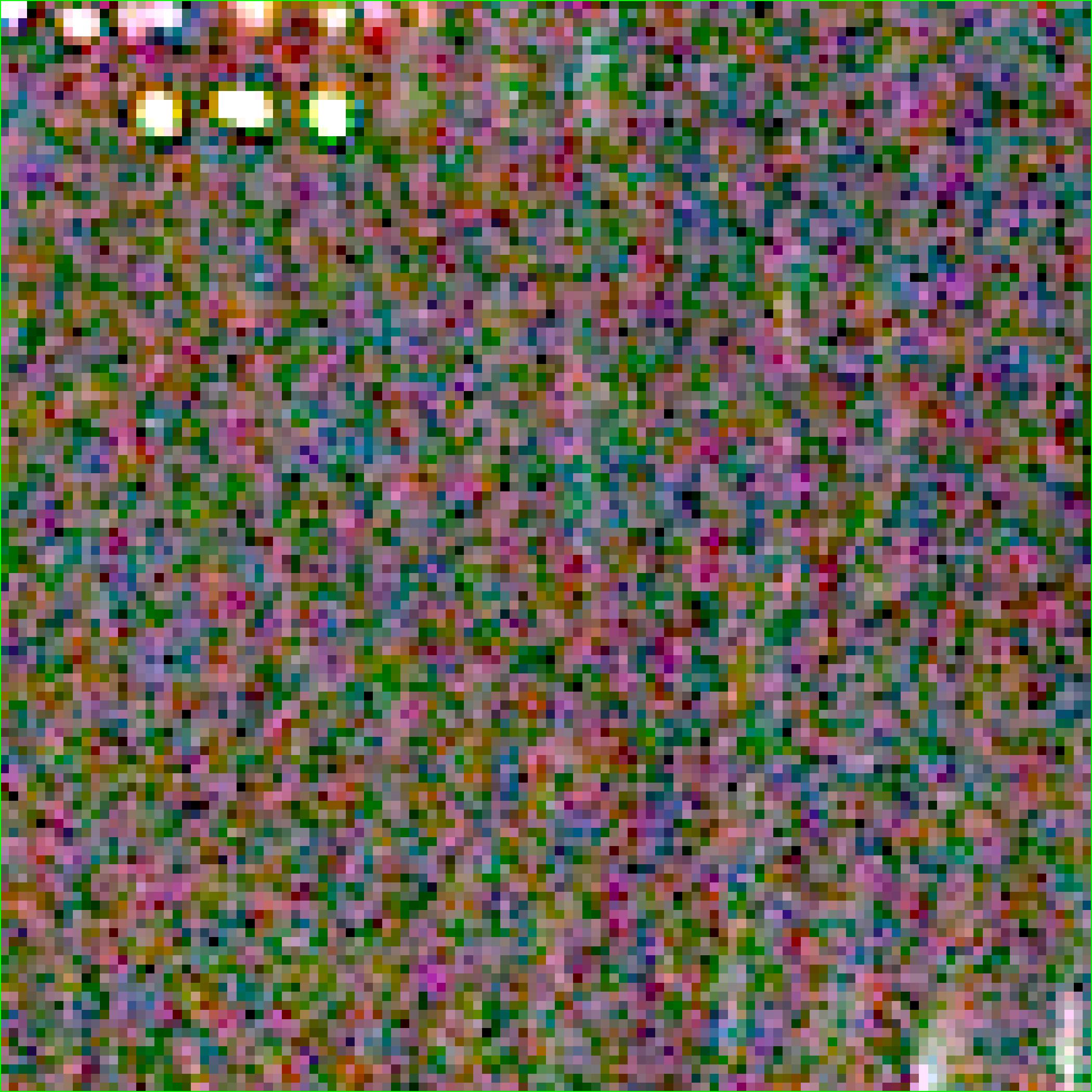}
    \end{minipage} 
    \begin{minipage}[b]{0.16\textwidth}
    \centering
    \includegraphics[width=1\textwidth]{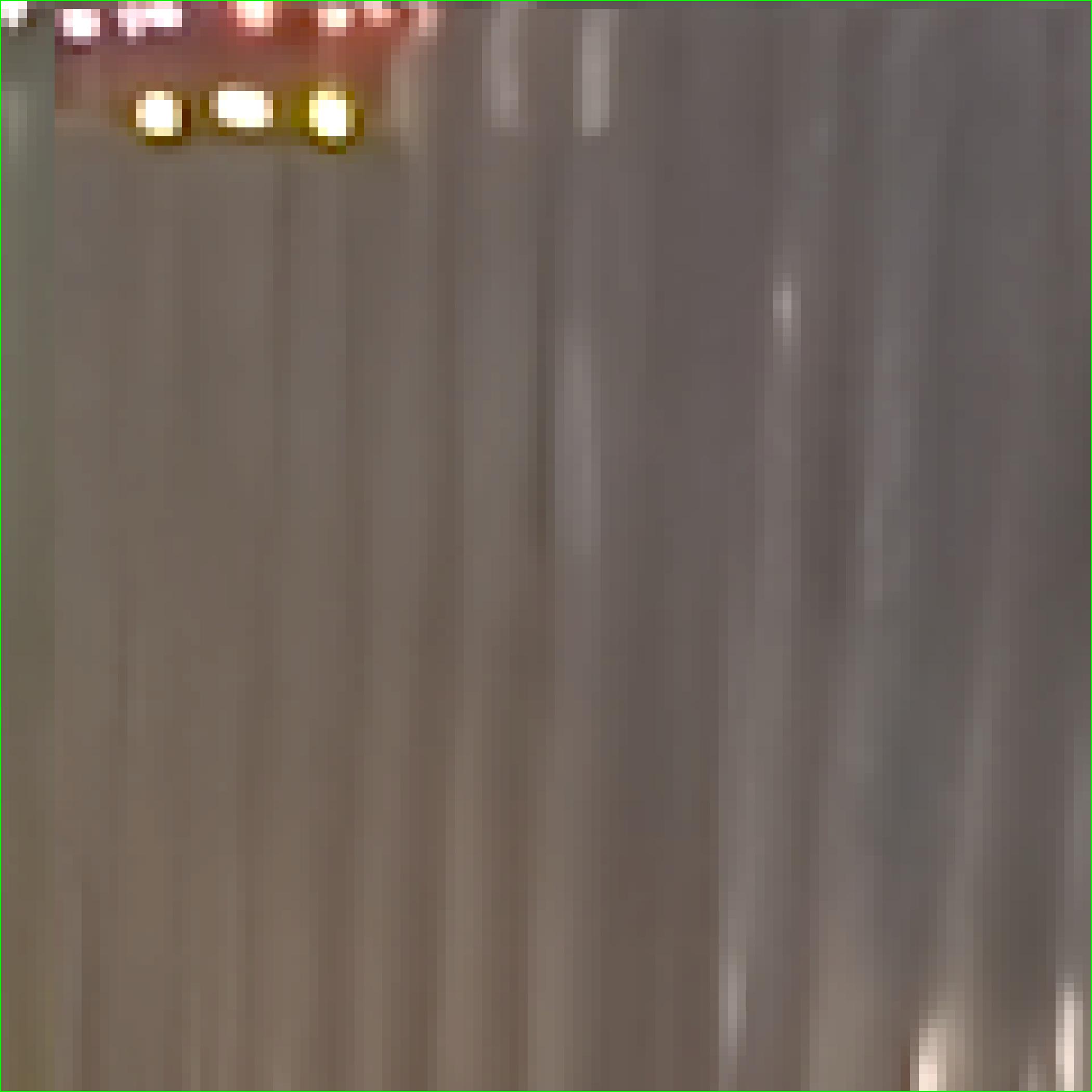}
    \end{minipage} 
    \begin{minipage}[b]{0.16\textwidth}
    \centering
    \includegraphics[width=1\textwidth]{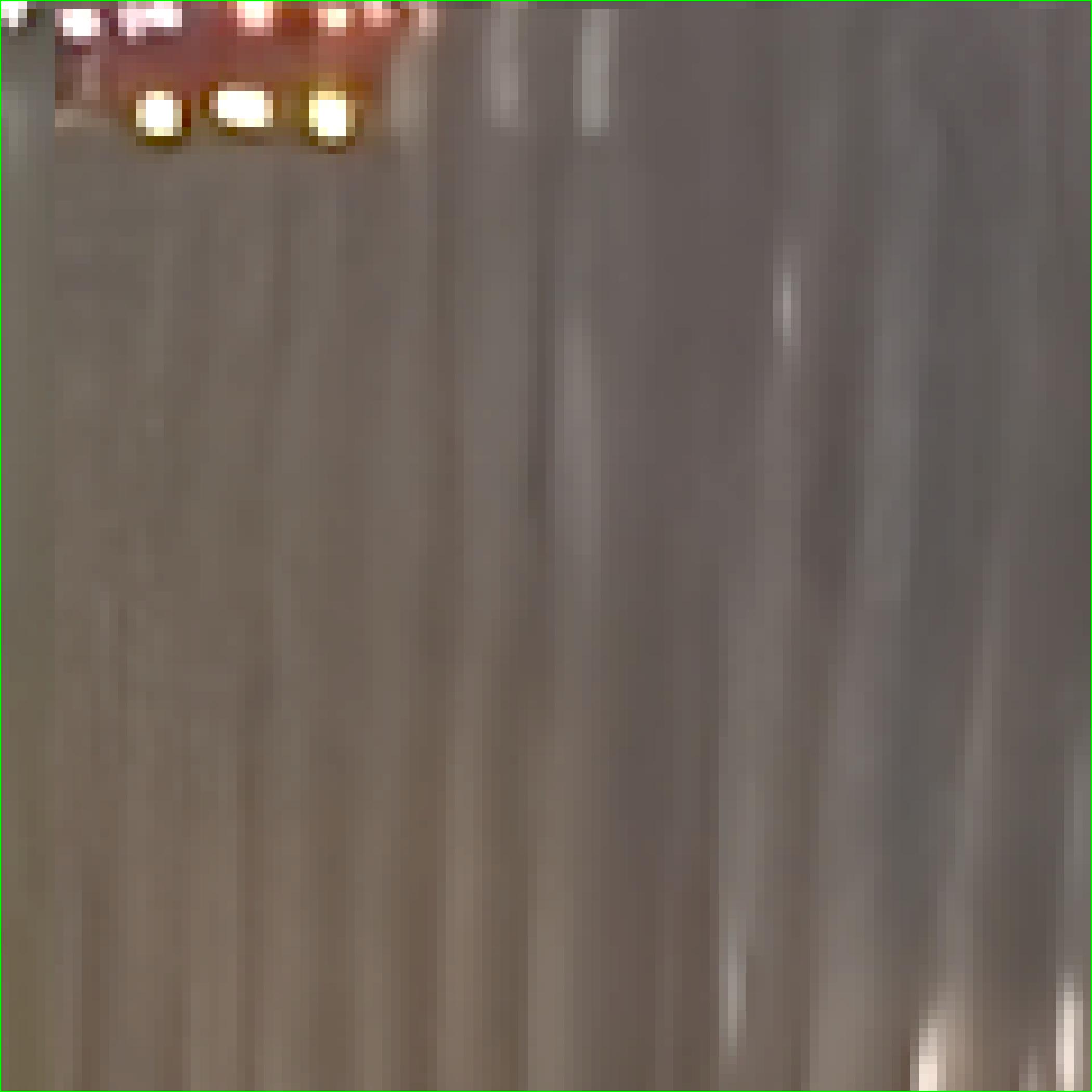}
    \end{minipage} 
    \begin{minipage}[b]{0.16\textwidth}
    \centering
    \includegraphics[width=1\textwidth]{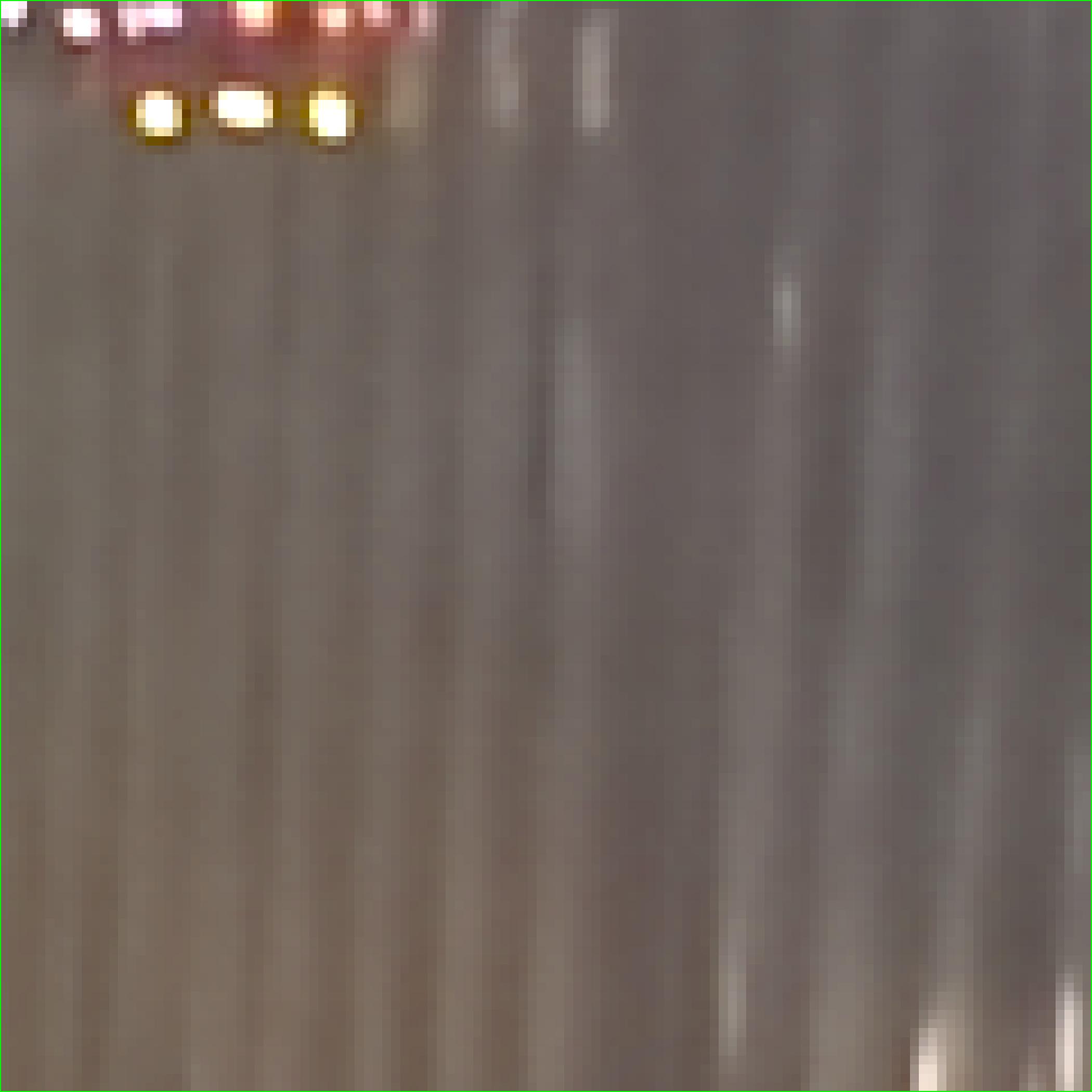}
    \end{minipage} 
    \begin{minipage}[b]{0.16\textwidth}
    \centering
    \includegraphics[width=1\textwidth]{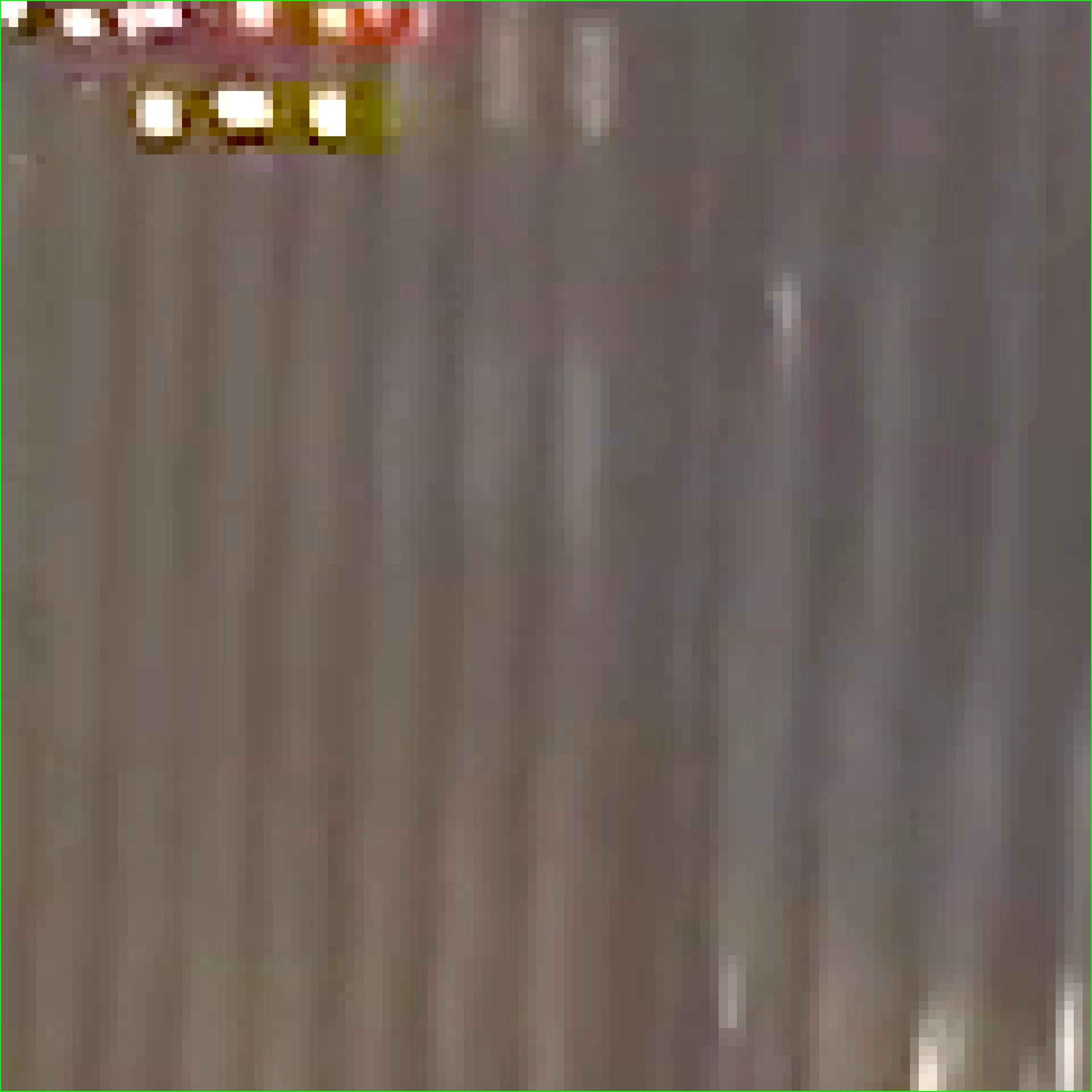}
    \end{minipage} 
\end{minipage}
\vspace{0.15cm}

\begin{minipage}[b]{1.0\textwidth}
\centering
    \begin{minipage}[b]{0.16\textwidth}
    \centering
    \includegraphics[width=1\textwidth]{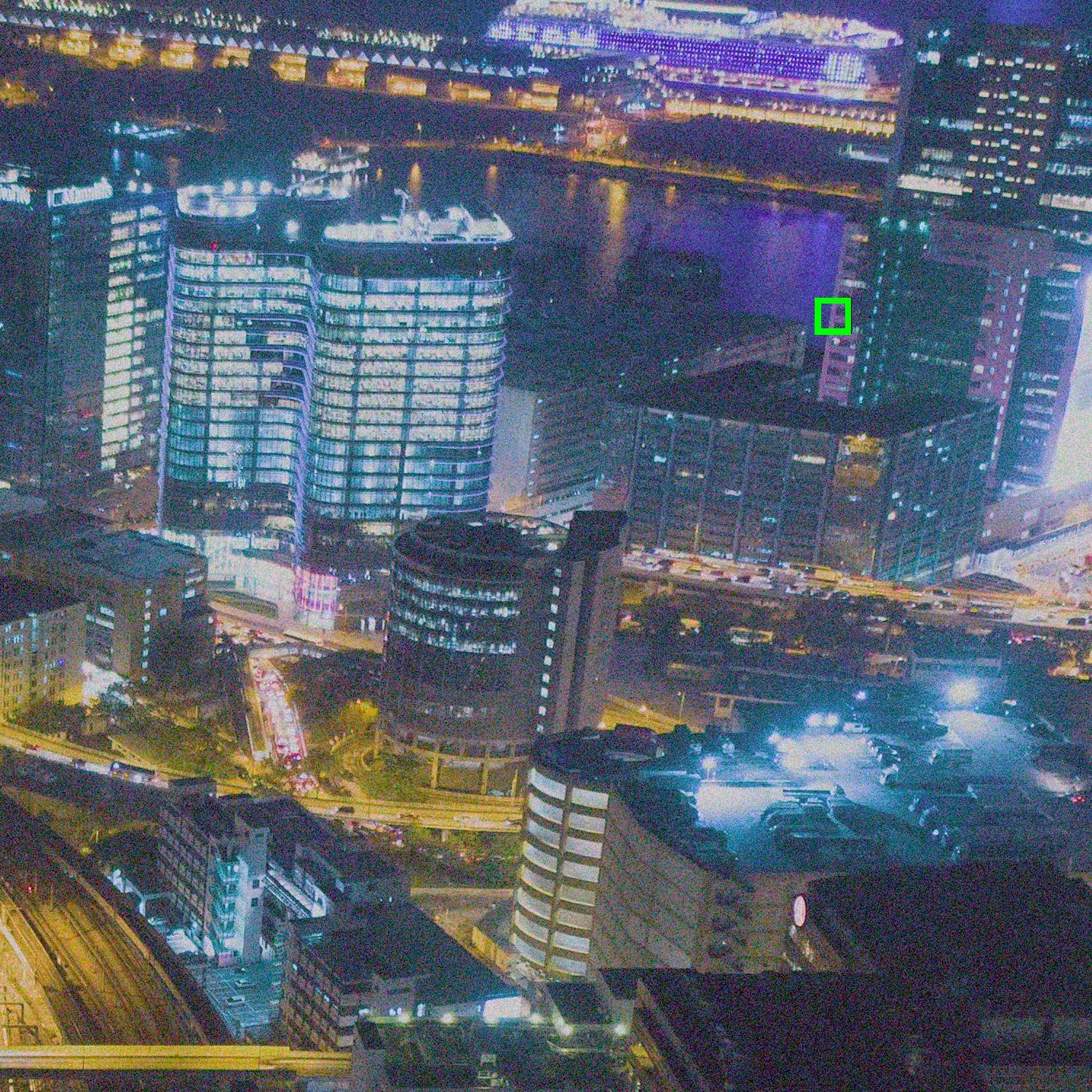}
    {\small (a) Noisy image}
    \end{minipage} 
    \begin{minipage}[b]{0.16\textwidth}
    \centering
    \includegraphics[width=1\textwidth]{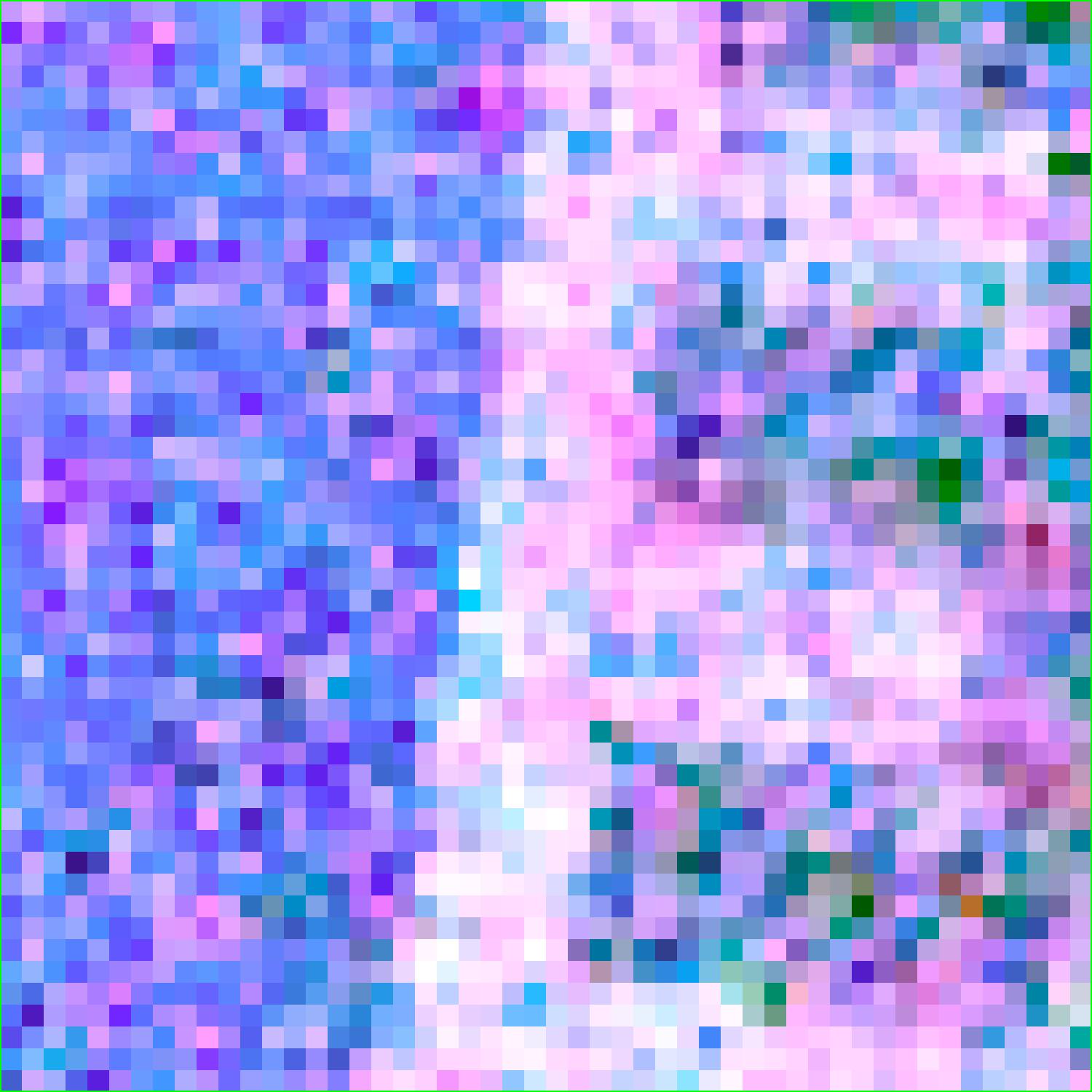}
    {\small (b) Noisy patch}
    \end{minipage} 
    \begin{minipage}[b]{0.16\textwidth}
    \centering
    \includegraphics[width=1\textwidth]{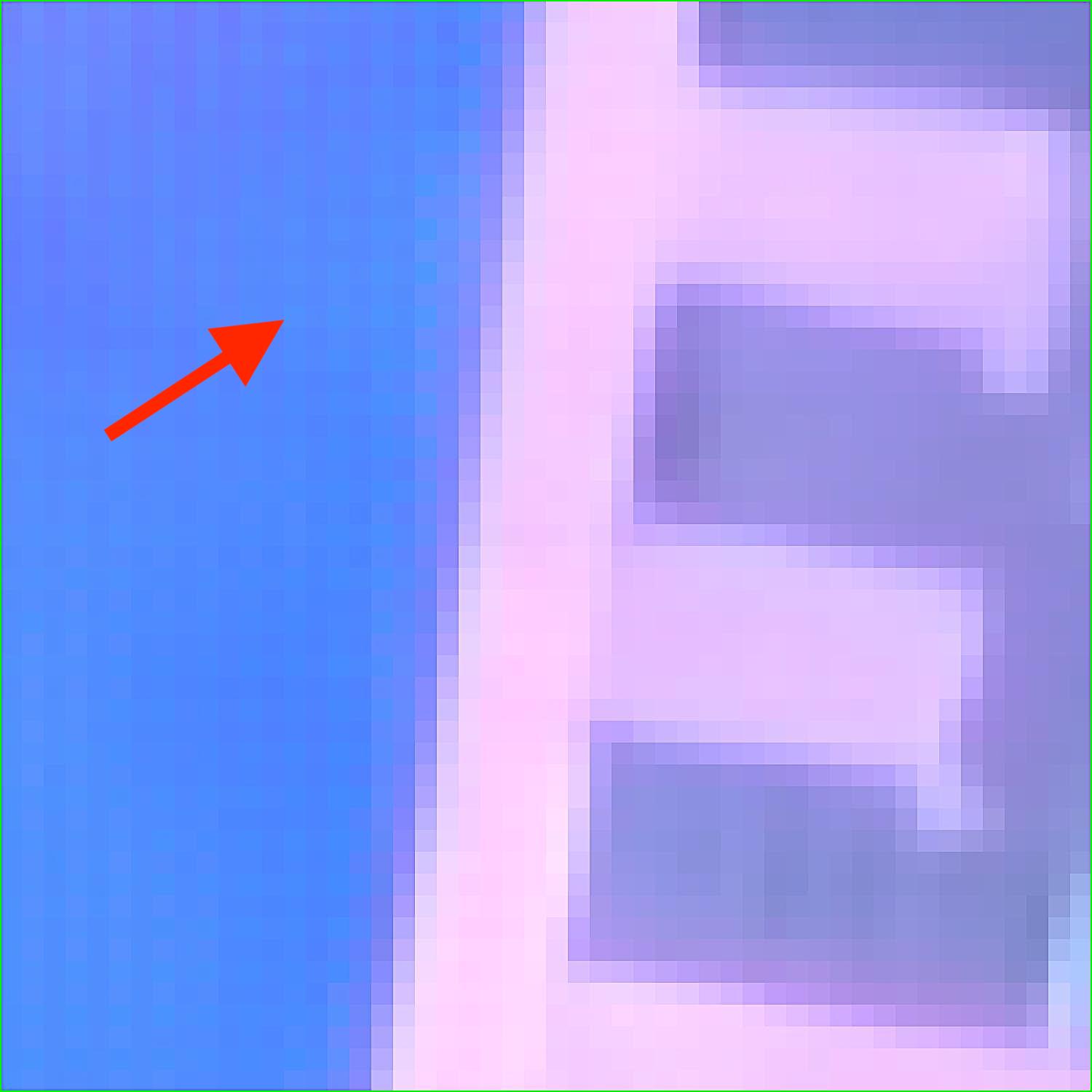}
    {\small (c) GCP-Net~\cite{guo2021joint}}
    \end{minipage} 
    \begin{minipage}[b]{0.16\textwidth}
    \centering
    \includegraphics[width=1\textwidth]{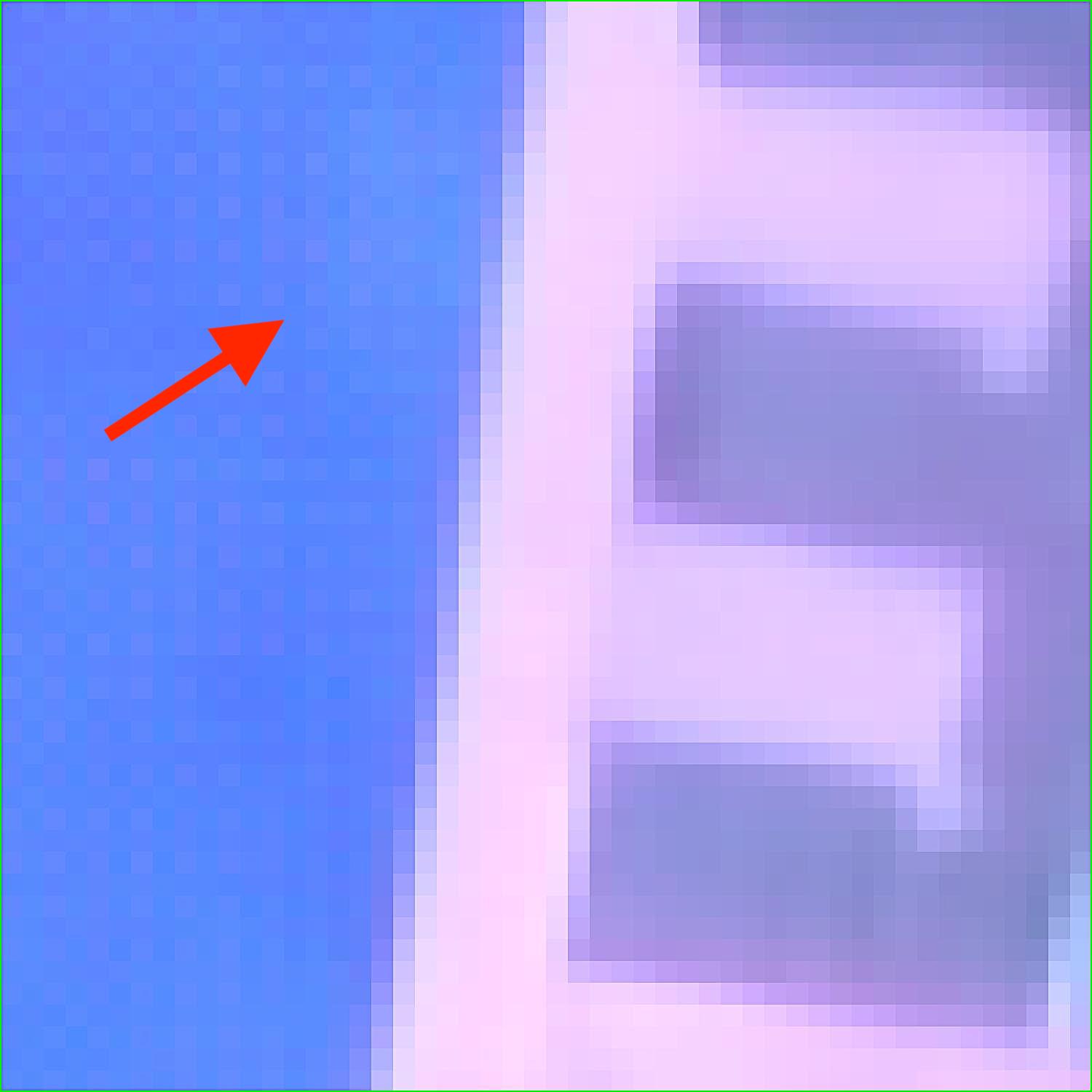}
    {\small (d) 2StageAlign~\cite{guo2022differentiable}}
    \end{minipage} 
    \begin{minipage}[b]{0.16\textwidth}
    \centering
    \includegraphics[width=1\textwidth]{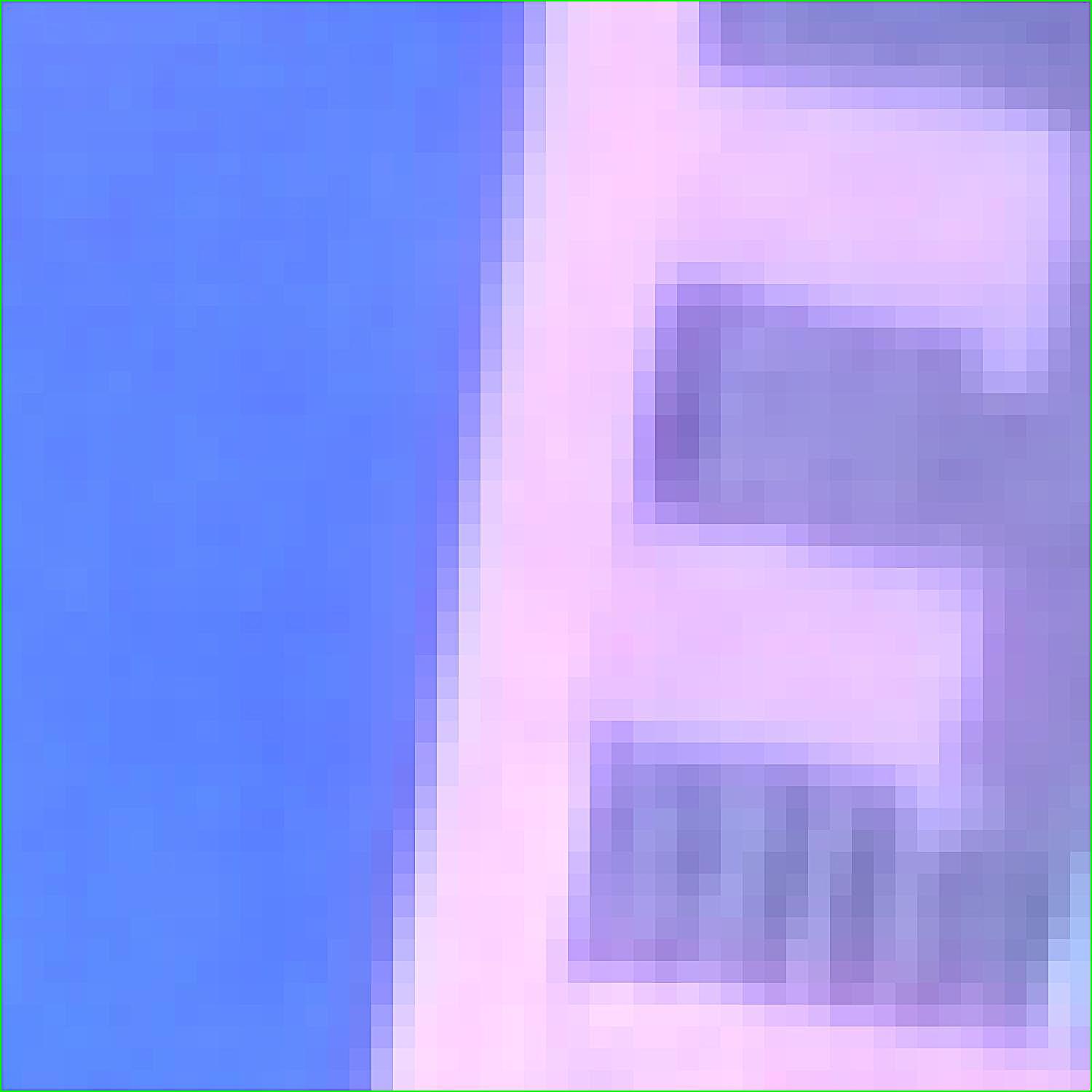}
    {\small (e) Ours}
    \end{minipage} 
    \begin{minipage}[b]{0.16\textwidth}
    \centering
    \includegraphics[width=1\textwidth]{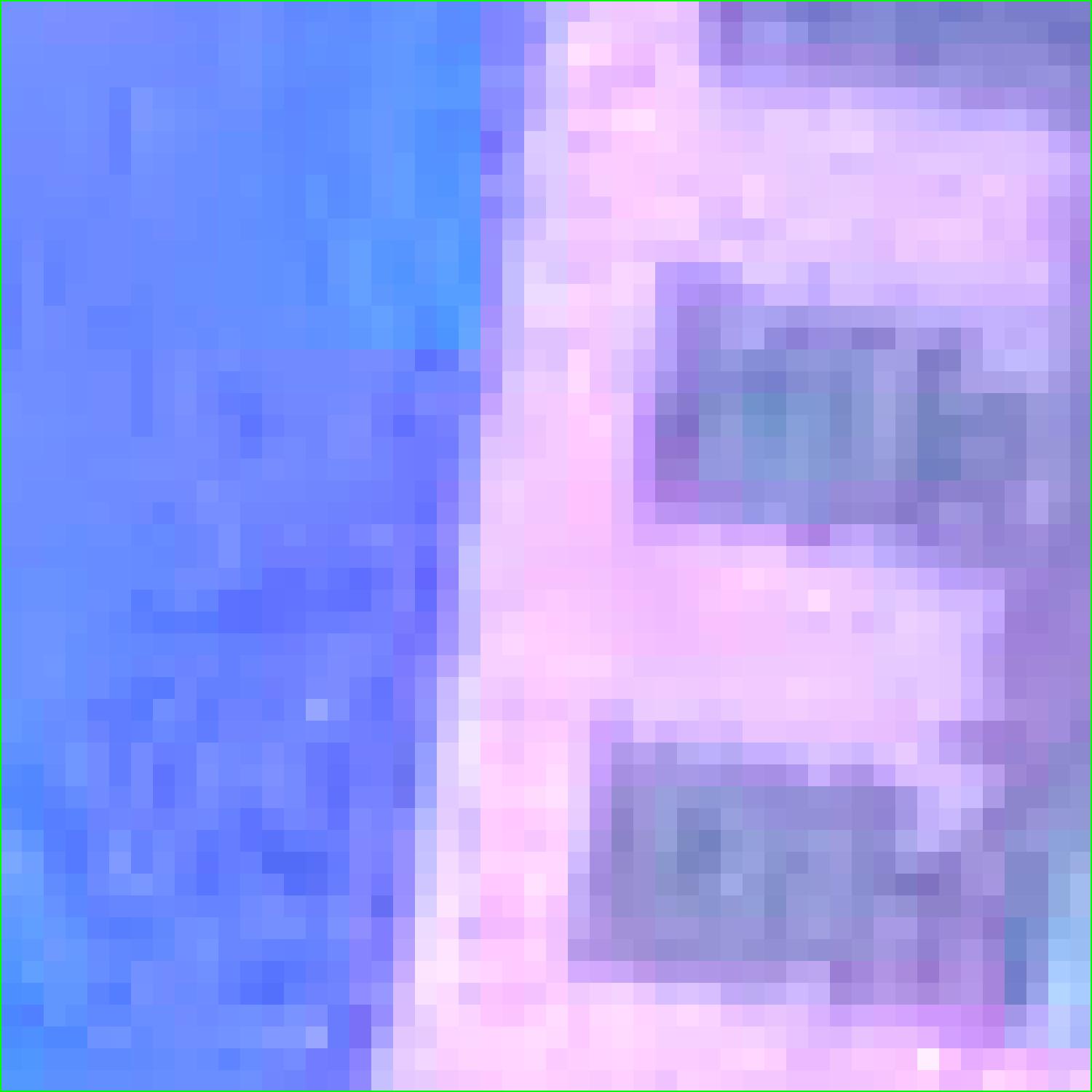}
    {\small (f) GT}
    \end{minipage} 
\end{minipage}
\caption{VJDD results by different methods on the 4KPix50 dataset. Zoom in to compare zipper artifacts from GCP-Net and 2StageAlign.}
\label{fig:4K}
\end{figure*}

\begin{figure*}[!h]
\centering
\begin{minipage}[b]{1.0\textwidth}
\centering
    \begin{minipage}[b]{0.24\textwidth}
    \centering
    \includegraphics[width=1\textwidth]{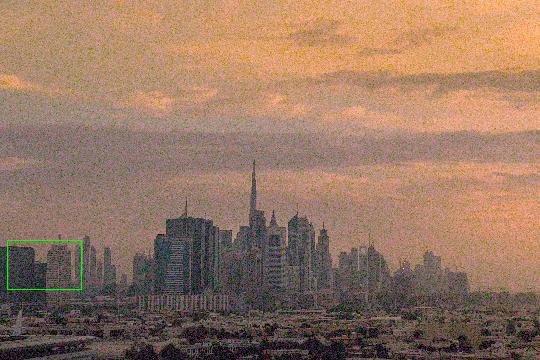}
    {\small Noise image}
    \end{minipage} 
    \begin{minipage}[b]{0.24\textwidth}
    \centering
    \includegraphics[width=1\textwidth]{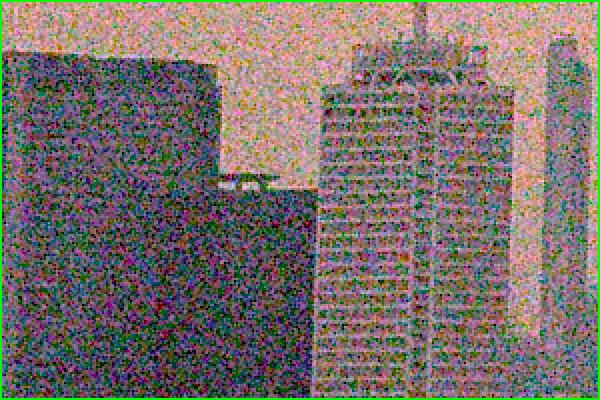}
    {\small Noise patch}
    \end{minipage} 
    \begin{minipage}[b]{0.24\textwidth}
    \centering
    \includegraphics[width=1\textwidth]{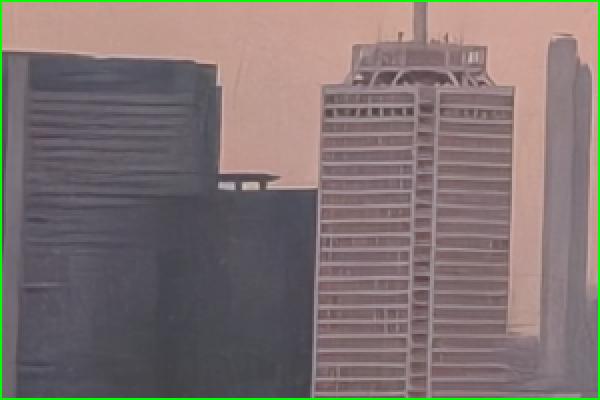}
    {\small EDVR-VJDD~\cite{wang2019edvr}}
    \end{minipage} 
    \begin{minipage}[b]{0.24\textwidth}
    \centering
    \includegraphics[width=1\textwidth]{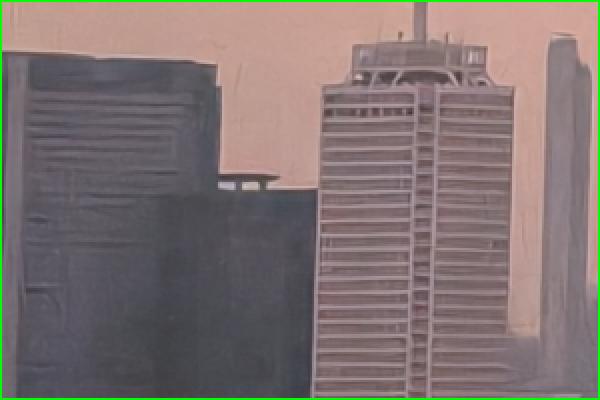}
    {\small RviDeNet-VJDD~\cite{yue2020supervised}}
    \end{minipage} 
\end{minipage}
\vspace{0.15cm}

\begin{minipage}[b]{1.0\textwidth}
\centering
    \begin{minipage}[b]{0.24\textwidth}
    \centering
    \includegraphics[width=1\textwidth]{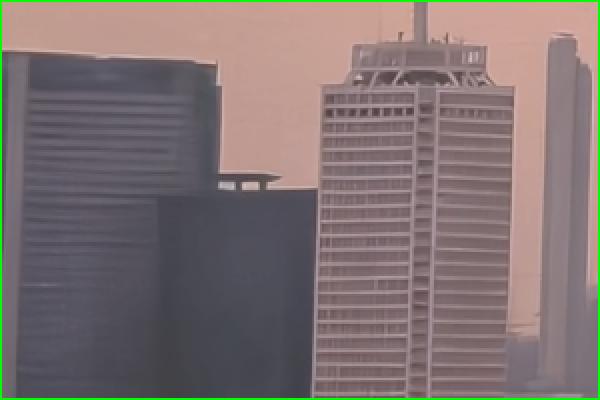}
    {\small GCP-Net~\cite{guo2021joint}}
    \end{minipage} 
    \begin{minipage}[b]{0.24\textwidth}
    \centering
    \includegraphics[width=1\textwidth]{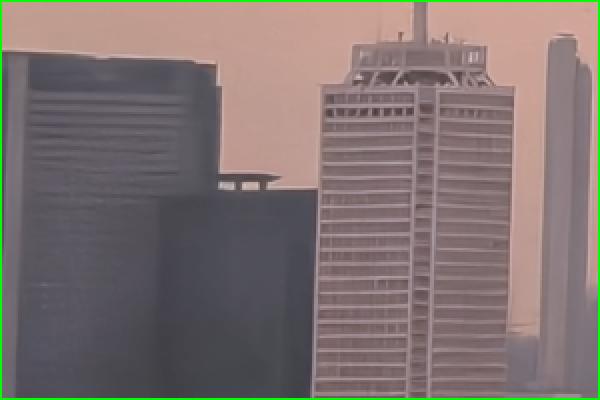}
    {\small 2StageAlign~\cite{guo2022differentiable}}
    \end{minipage} 
    \begin{minipage}[b]{0.24\textwidth}
    \centering
    \includegraphics[width=1\textwidth]{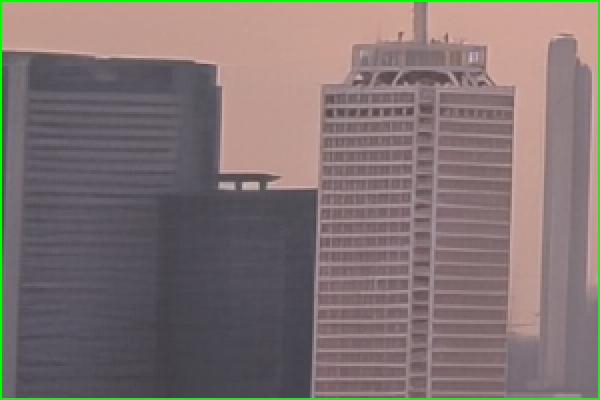}
    {\small Ours}
    \end{minipage} 
    \begin{minipage}[b]{0.24\textwidth}
    \centering
    \includegraphics[width=1\textwidth]{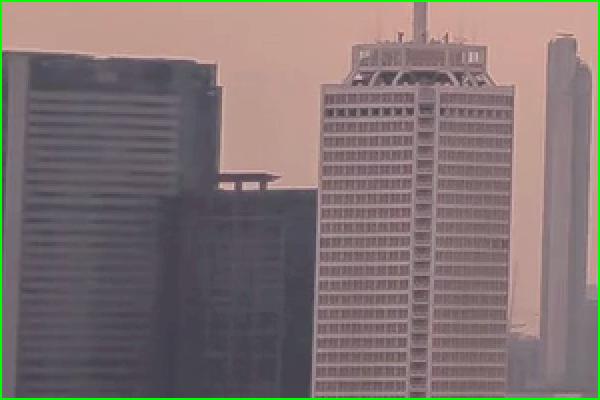}
    {\small GT}
    \end{minipage} 
\end{minipage}\vspace{0.15cm}

\begin{minipage}[b]{1.0\textwidth}
\centering
    \begin{minipage}[b]{0.24\textwidth}
    \centering
    \includegraphics[width=1\textwidth]{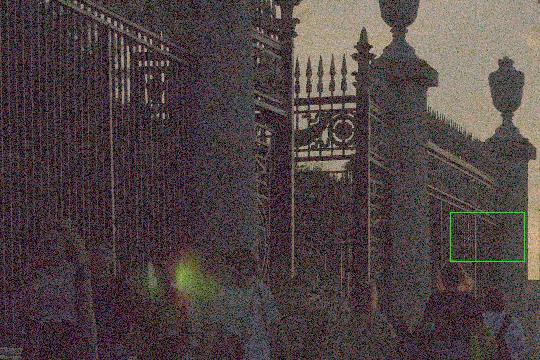}
    {\small Noise image}
    \end{minipage} 
    \begin{minipage}[b]{0.24\textwidth}
    \centering
    \includegraphics[width=1\textwidth]{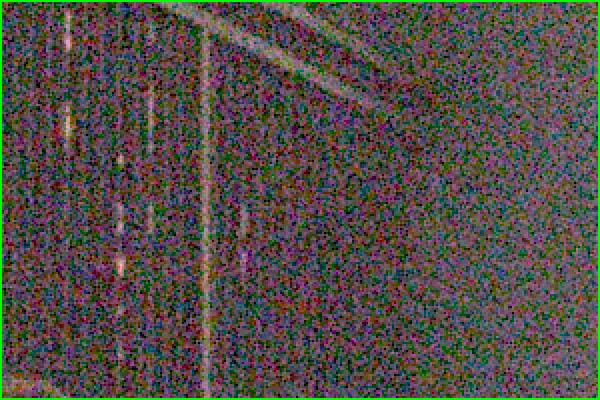}
    {\small Noise patch}
    \end{minipage} 
    \begin{minipage}[b]{0.24\textwidth}
    \centering
    \includegraphics[width=1\textwidth]{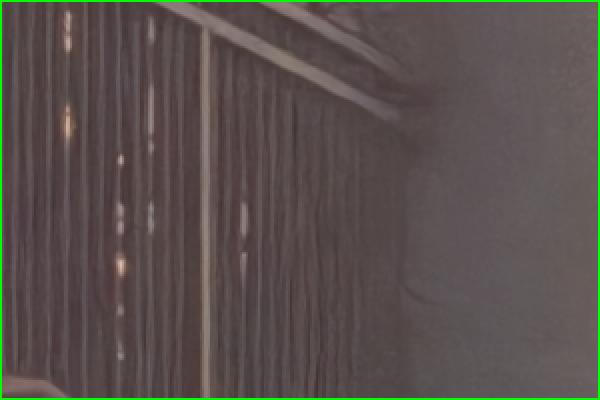}
    {\small EDVR-VJDD~\cite{wang2019edvr}}
    \end{minipage} 
    \begin{minipage}[b]{0.24\textwidth}
    \centering
    \includegraphics[width=1\textwidth]{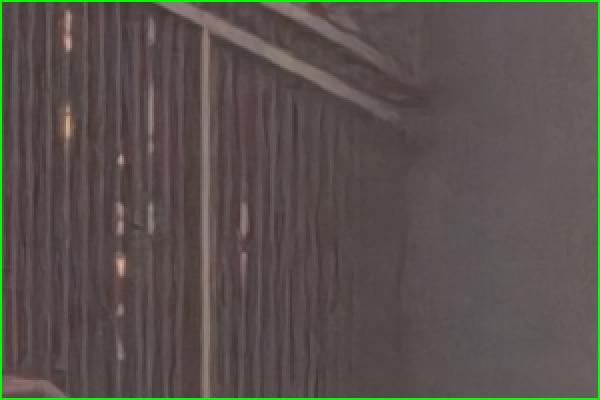}
    {\small RviDeNet-VJDD~\cite{yue2020supervised}}
    \end{minipage} 
\end{minipage}
\vspace{0.15cm}

\begin{minipage}[b]{1.0\textwidth}
\centering
    \begin{minipage}[b]{0.24\textwidth}
    \centering
    \includegraphics[width=1\textwidth]{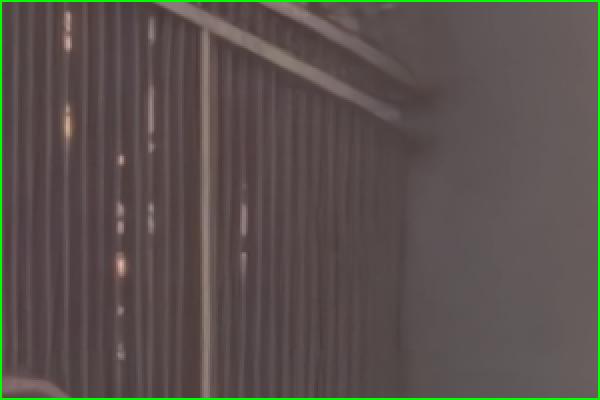}
    {\small GCP-Net~\cite{guo2021joint}}
    \end{minipage} 
    \begin{minipage}[b]{0.24\textwidth}
    \centering
    \includegraphics[width=1\textwidth]{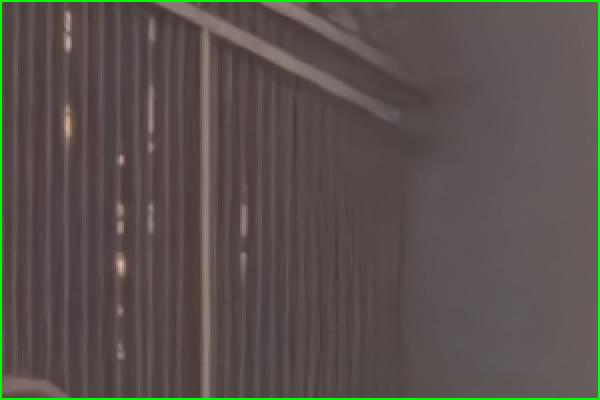}
    {\small 2StageAlign~\cite{guo2022differentiable}}
    \end{minipage} 
    \begin{minipage}[b]{0.24\textwidth}
    \centering
    \includegraphics[width=1\textwidth]{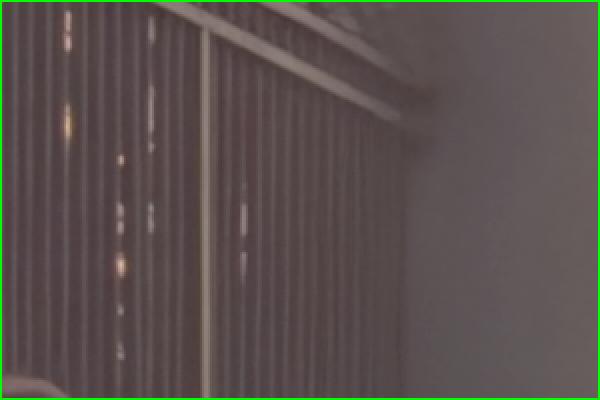}
    {\small Ours}
    \end{minipage} 
    \begin{minipage}[b]{0.24\textwidth}
    \centering
    \includegraphics[width=1\textwidth]{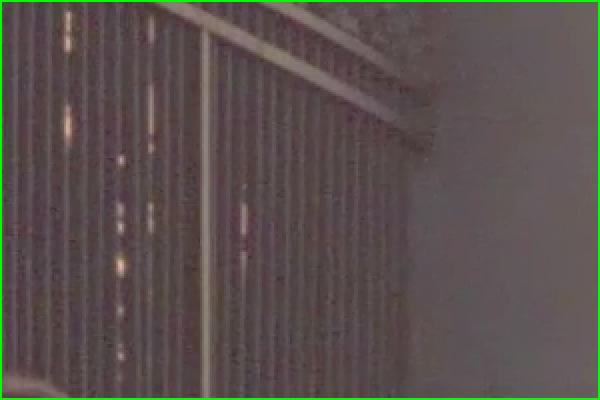}
    {\small GT}
    \end{minipage} 
\end{minipage}
\caption{VJDD results by different methods on 4KPix50 dataset.}
\label{fig:supp_4kpix_1}
\end{figure*}

\textbf{Test Dataset and Competing Methods.} We evaluate the performance of our method on two datasets, \ie, REDS4~\cite{wang2019edvr} and our collected dataset, named 4KPix50. REDS4 is a commonly used test set for video restoration tasks; however, it has only 4 sequences and the resolution is only 720p. To more comprehensively evaluate the VJDD methods, we collected 50 videos of 4K resolution from the Pixabay website~\cite{pixabeydataset}, building the 4KPix50 dataset, which contains a diverse range of scenes, including indoor scenes, outdoor scenes, landscapes, human subjects, nighttime footage, animals, \etc, as shown in \cref{fig:all_4K}. The motion statistics in 4KPix50 are also illustrated in \cref{fig:all_4K}. One can see that it is a long-tailed distribution and there are many large motions bigger than 20 pixels. 

We compare our method against state-of-the-art multiframe JDD methods, including GCP-Net~\cite{guo2021joint} and 2StageAlign~\cite{guo2022differentiable}, as well as the leading burst denoising algorithms, including EDVR~\cite{wang2019edvr} and RviDeNet~\cite{yue2020supervised}. Note that we re-train EDVR and RviDeNet following the experimental setup in~\cite{guo2021joint,guo2022differentiable} to adapt them on the VJDD task, denoted as EDVR-VJDD and RviDeNet-VJDD.

\subsection{Results on Restoration Accuracy}
The quantitative results of competing VJDD methods on the REDS4 and 4KPix50 datasets are presented in \cref{table:reds4}. We employ the restoration fidelity metrics PSNR and SSIM, and the perceptual quality metrics LPIPS~\cite{zhang2018unreasonable} and DISTS~\cite{ding2020image} for evaluation. To more comprehensively evaluate the performance of VJDD methods, we present the results under both high and low noise levels. One can see that our proposed method achieves the best results under both the two datasets and the two noise levels in terms of all evaluation metrics (\ie, PSNR/SSIM/LPIPS/DISTS). Specifically, by leveraging the proposed latent space propagation strategy, our method achieves 0.37dB and 0.6dB improvement over the previous leading method 2StageAlign on REDS4 and 4KPix20, respectively, under the high noise level. By using the perceptual loss in the training process, our method also achieves impressive improvement on perceptual quality metrics over 2StageAlign, \ie, improving LPIPS from 0.1257 to 0.0977 (about 22\%) and DISTS from 0.0757 to 0.0474 (about 37\%) on the REDS4 dataset. 

\cref{fig:Reds4,fig:4K,fig:supp_4kpix_1} show the qualitative comparisons of competing methods on REDS4 and 4KPix20, respectively. One can see that, due to the integration of perceptual loss, our approach facilitates the restoration of intricate details. Specifically, from the comparison of stone textures in \cref{fig:Reds4} and the comparison of structural features in \cref{fig:4K}, it is evident that our method preserves a richer amount of textures than the previous SOTA methods GCP-Net and 2StageAlign. Meanwhile, the introduction of temporal stability constraints (\ie, $\mathcal{L}_{DTC}$ and $\mathcal{L}_{RPC}$) ensures that the details restored by our method are visually pleasing and temporally consistent.

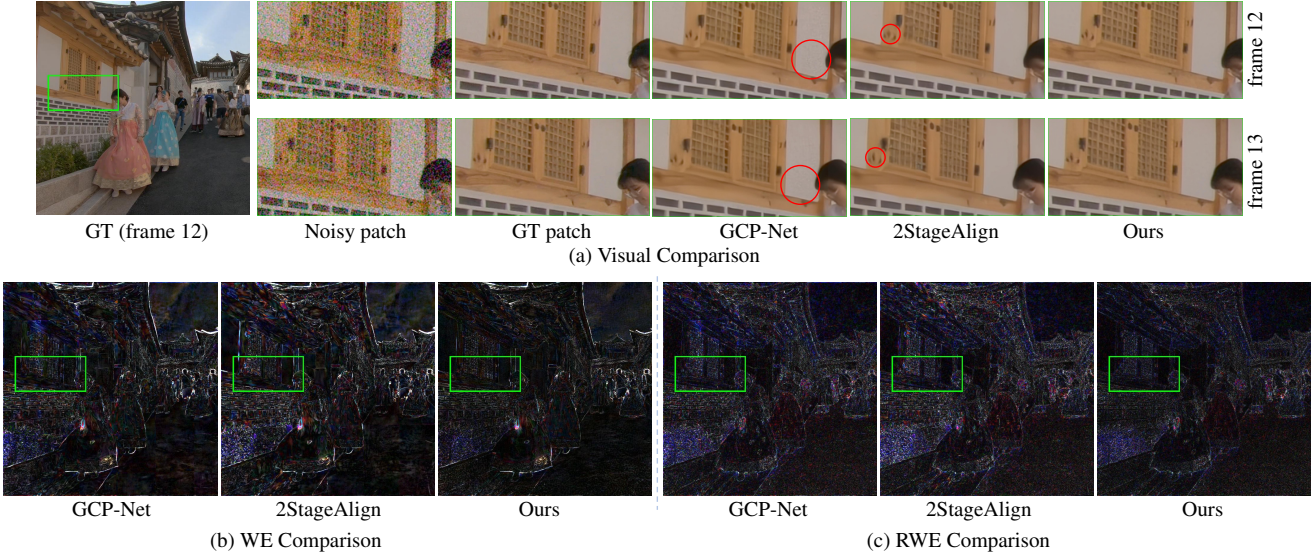
\begin{figure*}[!tbp]
\centering
\begin{overpic}[width=1.0\textwidth]{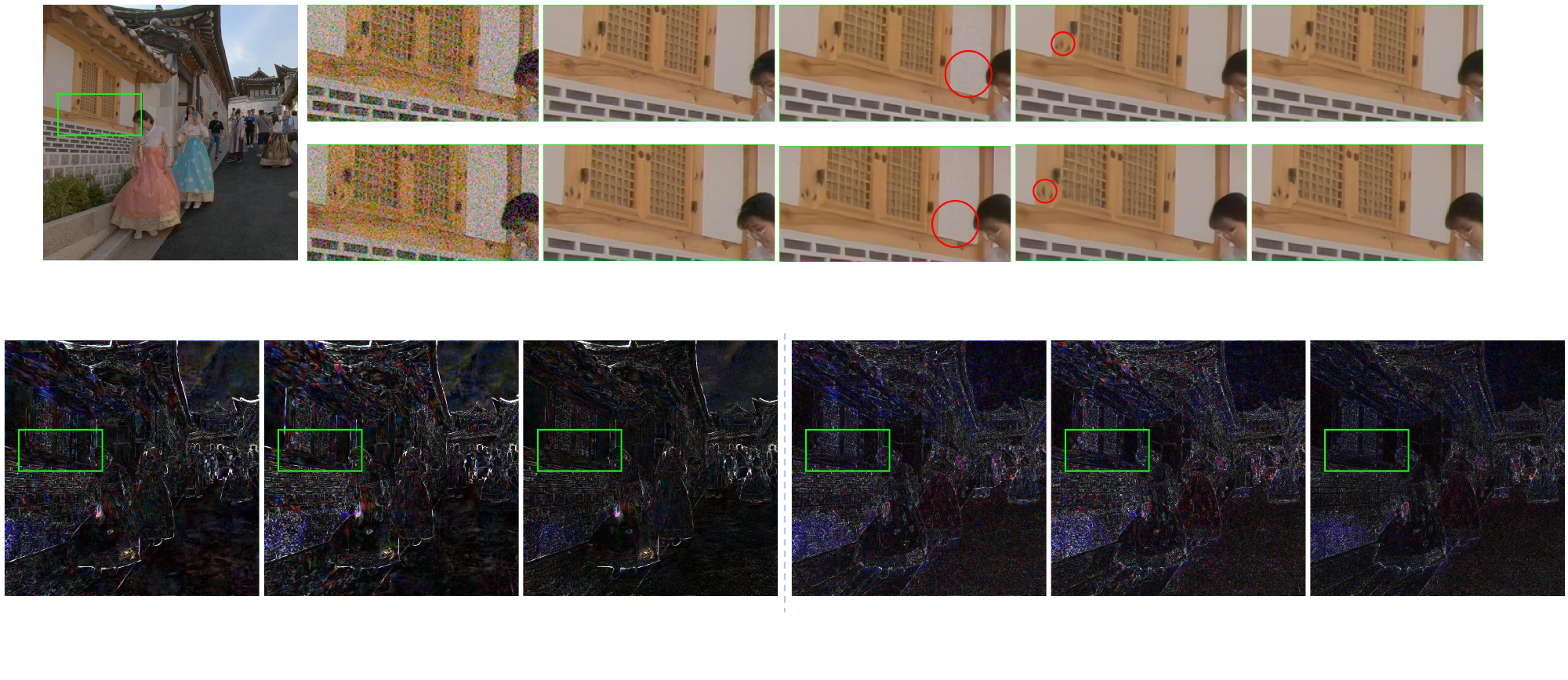}
    \put(43.3,23.2){\footnotesize (a) Visual Comparison}
    \put(6.5,25.0){\footnotesize GT (frame 12)}
    \put(23.2,25.0){\footnotesize Noisy patch}
    \put(39,25.0){\footnotesize GT patch}
    \put(54.8,25.0){\footnotesize GCP-Net}
    \put(68,25.0){\footnotesize 2StageAlign}
    \put(85.5,25.0){\footnotesize Ours}
    \put(95,36.5){\footnotesize \rotatebox{90}{frame 12}}
    \put(95,27.2){\footnotesize \rotatebox{90}{frame 13}}

    \put(5.5,3.8){\footnotesize GCP-Net}
    \put(21,3.8){\footnotesize 2StageAlign}
    \put(39.5,3.8){\footnotesize Ours}
    \put(16,1.5){\footnotesize (b) WE Comparison}

    \put(55.5,3.8){\footnotesize GCP-Net}
    \put(70.5,3.8){\footnotesize 2StageAlign}
    \put(90,3.8){\footnotesize Ours}
    \put(66,1.5){\footnotesize (c) RWE Comparison}
    
\end{overpic}\vspace{-0.4cm}
\caption{Temporal visual comparison of VJDD results by different methods on the $12$-th and $13$-th frames of Clip020 in REDS4 dataset. Note that the visualization of wrapping error (WE) and relational wrapping error (RWE) are amplified by 10 times for better observation. Better viewed with zoom-in on screen.}
\label{fig:tc_compare}
\vspace{-0.2cm}
\end{figure*}

\subsection{Results on Temporal Consistency}
We quantitatively evaluate the temporal consistency using the widely used metrics, including warping error (WE)~\cite{chang2019free,wang2020consistent,dai2022video} and tOF~\cite{chu2020learning}, as well as a metric proposed by us, \ie, relational warping error (RWE). The WE models the mean absolute error (MAE) between the aligned adjacent output frames and the tOF models the MAE between the optical flow estimated from restored frames and ground-truth frames. However, due to the natural intensity difference between frames and the estimation error of optical flow, the WE is not zero even for ground-truth videos and hence it cannot reflect the temporal consistency accurately. Therefore, we design an improved metric RWE as follows:
\begin{equation}
\small
    \text{RWE}(\hat{x}_{t+1}, \hat{x}_{t}) = \Vert M((\hat{x}_{t+1}^{t} -\hat{x}_{t}) - (x_{t+1}^{t} - x_{t})) \Vert, 
\end{equation}
where $\hat{x}_{t+1}^{t} = \tau(\hat{x}_{t+1}, \hat{f}_{t+1\rightarrow t})$, and $\hat{f}_{t+1\rightarrow t}$ and $M$ are the estimated optical flow and occlusion map by SpyNet~\cite{ranjan2017optical}. RWE subtracts the difference between adjacent ground-truth $(x_{t+1}^{t} - x_{t})$, which considers the natural variations of video frames and the estimation error of optical flow, and hence can more accurately model the temporal consistency.

The quantitative results are shown in \cref{table:reds4}. One can see that our model obtains the most stable results in terms of WE, tOF and RWE on images with high noise level. The qualitative comparison of different methods on REDS4 is shown in \cref{fig:tc_compare}. From \cref{fig:tc_compare} (a), we can see that GCP-Net retains noise, resulting in temporal jittering in flat regions of consecutive frames. For 2StageAlign, the temporal jittering happens in many structural regions of the object across different frames. Our proposed method leverages the consistent latent space propagation strategy and temporal regularizations, preserving effectively the temporal consistency in both flat and texture regions. The visual comparisons of WE and RWE in \cref{fig:tc_compare} (b) and (c) further validate the effectiveness of our approach in achieving temporally consistent VJDD results.

\begin{table}[!tbp]
\centering
\caption{Comparison of different variants on the REDS4 dataset.} 
\begin{tabular}{c c c c}
\toprule
Methods  &LPIPS &WE ($10^{-2}$) &RWE ($10^{-2}$)\\
\midrule
w/o $\mathcal{L}_t$, w/o $\mathcal{L}_{p}$ &0.1205 &1.23 &1.16 \\
w/o $\mathcal{L}_t$ &0.0960 &1.32 &1.23 \\
\midrule
w/o $\mathcal{L}_t$, w/o $H_t$ &0.1007 &1.35 &1.25 \\
w/o $\mathcal{L}_t$, w/ $SH_t$ &0.0982 &1.33 &1.23 \\
w/o $\mathcal{L}_{RPC}$ &0.0968 &1.26 &1.19 \\
w/o $\mathcal{L}_{DTC}$ &0.0963 &1.28 &1.21 \\
\midrule
full model &0.0977 &1.23 &1.17 \\

\toprule
\end{tabular} 
\label{table:ablation}
\end{table}

\subsection{Ablation Study}
We first evaluate the impact of perceptual loss ($\mathcal{L}_{p}$) and temporal losses ($\mathcal{L}_{DTC}$, $\mathcal{L}_{DTC-l}$, $\mathcal{L}_{RPC}$) on the VJDD results. For the convenience of expression, we denote by $\mathcal{L}_{t}$ all the three temporal losses. The results are shown in \cref{table:ablation}. One can see that the use of $\mathcal{L}_{p}$ significantly enhances the perceptual quality of the VJDD outputs, as validated by the smaller LPIPS scores of model ``w/o $\mathcal{L}_t$" in comparison with the model ``w/o $\mathcal{L}_t$, w/o $\mathcal{L}_{p}$". However, employing only $\mathcal{L}_{p}$ introduces noticeable temporal inconsistency, as observed by the increase in WE from $1.23\times 10^{-2}$ to $1.32\times 10^{-2}$ and RWE from $1.16\times 10^{-2}$ to $1.23\times 10^{-2}$. In contrast, by incorporating the proposed temporal losses $\mathcal{L}_{t}$, our full model achieves improved perceptual quality while maintaining temporal consistency. 

Secondly, we assess the impact of our propagation scheme on  temporal consistency. We train four more variants: without utilizing temporal losses and temporal hidden state (w/o $\mathcal{L}_t$, w/o $H_t$), without using temporal losses but using shallow feature as the hidden state (w/o $\mathcal{L}_t$, w/ $SH_t$), without using DTC loss (w/o $\mathcal{L}_{DTC}$), and without using RPC loss (w/o $\mathcal{L}_{RPC}$). The results in \cref{table:ablation} clearly indicate that both the consistent latent space propagation and the temporal losses contribute favorably to enhancing the temporal consistency of the VJDD results. Our full model achieves an excellent trade-off between perceptual quality and temporal consistency.

\section{Conclusion}
We investigated the problem of video joint denoising and demosaicking (VJDD) and proposed a novel approach that achieved new state-of-the-arts in both restoration accuracy and temporal consistency. Our network adopted a consistent latent space propagation strategy, which utilized the restoration features of previous frames as prior knowledge for recovering the current frame. Moreover, we introduced two novel temporal losses: the data temporal consistency (DTC) loss and the relational perception consistency (RPC) loss. The DTC loss utilized synthetic videos with ground-truth transformations to endow consistency among consecutive frames, avoiding the error accumulation problem in flow-based methods. The RPC loss encouraged the perceptual relations of predicted frames to align with those of the ground-truth, ensuring high quality restoration without over-smoothing details. Experimental results demonstrated the significant improvements of our method in restoration accuracy, perceptual quality and temporal consistency over existing VJDD methods.

\vspace{4mm}
\noindent\textbf{Acknowledgements} This work is supported by the Hong Kong RGC RIF grant (R5001-18) and the PolyU-OPPO Joint Innovation Lab.

{
    \small
    \bibliographystyle{ieeenat_fullname}
    \bibliography{main}

\begin{thebibliography}{62}
\providecommand{\natexlab}[1]{#1}
\providecommand{\url}[1]{\texttt{#1}}
\expandafter\ifx\csname urlstyle\endcsname\relax
  \providecommand{\doi}[1]{doi: #1}\else
  \providecommand{\doi}{doi: \begingroup \urlstyle{rm}\Url}\fi

\bibitem[Brooks et~al.(2019)Brooks, Mildenhall, Xue, Chen, Sharlet, and Barron]{brooks2019unprocessing}
Tim Brooks, Ben Mildenhall, Tianfan Xue, Jiawen Chen, Dillon Sharlet, and Jonathan~T Barron.
\newblock Unprocessing images for learned raw denoising.
\newblock In \emph{IEEE Conf. Comput. Vis. Pattern Recog.}, pages 11036--11045, 2019.

\bibitem[Chan et~al.(2020)Chan, Wang, Yu, Dong, and Loy]{chan2020basicvsr}
Kelvin~CK Chan, Xintao Wang, Ke Yu, Chao Dong, and Chen~Change Loy.
\newblock Basicvsr: The search for essential components in video super-resolution and beyond.
\newblock \emph{arXiv preprint arXiv:2012.02181}, 2020.

\bibitem[Chan et~al.(2021)Chan, Wang, Yu, Dong, and Loy]{chan2021understanding}
Kelvin~CK Chan, Xintao Wang, Ke Yu, Chao Dong, and Chen~Change Loy.
\newblock Understanding deformable alignment in video super-resolution.
\newblock In \emph{AAAI}, pages 973--981, 2021.

\bibitem[Chan et~al.(2022{\natexlab{a}})Chan, Zhou, Xu, and Loy]{chan2022basicvsr++}
Kelvin~CK Chan, Shangchen Zhou, Xiangyu Xu, and Chen~Change Loy.
\newblock Basicvsr++: Improving video super-resolution with enhanced propagation and alignment.
\newblock In \emph{IEEE Conf. Comput. Vis. Pattern Recog.}, pages 5972--5981, 2022{\natexlab{a}}.

\bibitem[Chan et~al.(2022{\natexlab{b}})Chan, Zhou, Xu, and Loy]{chan2022investigating}
Kelvin~CK Chan, Shangchen Zhou, Xiangyu Xu, and Chen~Change Loy.
\newblock Investigating tradeoffs in real-world video super-resolution.
\newblock In \emph{IEEE Conf. Comput. Vis. Pattern Recog.}, pages 5962--5971, 2022{\natexlab{b}}.

\bibitem[Chang et~al.(2019)Chang, Liu, Lee, and Hsu]{chang2019free}
Ya-Liang Chang, Zhe~Yu Liu, Kuan-Ying Lee, and Winston Hsu.
\newblock Free-form video inpainting with 3d gated convolution and temporal patchgan.
\newblock In \emph{Int. Conf. Comput. Vis.}, pages 9066--9075, 2019.

\bibitem[Chu et~al.(2020)Chu, Xie, Mayer, Leal-Taix{\'e}, and Thuerey]{chu2020learning}
Mengyu Chu, You Xie, Jonas Mayer, Laura Leal-Taix{\'e}, and Nils Thuerey.
\newblock Learning temporal coherence via self-supervision for gan-based video generation.
\newblock \emph{ACM Trans. Graph.}, 39\penalty0 (4):\penalty0 75--1, 2020.

\bibitem[Cok(1987)]{cok1987signal}
David~R Cok.
\newblock Signal processing method and apparatus for producing interpolated chrominance values in a sampled color image signal, 1987.
\newblock US Patent 4,642,678.

\bibitem[Condat and Mosaddegh(2012)]{condat2012joint}
Laurent Condat and Saleh Mosaddegh.
\newblock Joint demosaicking and denoising by total variation minimization.
\newblock In \emph{IEEE Int. Conf. Image Process.}, pages 2781--2784. IEEE, 2012.

\bibitem[Dai et~al.(2022)Dai, Yu, Ma, Zhang, Li, Li, Shen, and Qi]{dai2022video}
Peng Dai, Xin Yu, Lan Ma, Baoheng Zhang, Jia Li, Wenbo Li, Jiajun Shen, and Xiaojuan Qi.
\newblock Video demoireing with relation-based temporal consistency.
\newblock In \emph{IEEE Conf. Comput. Vis. Pattern Recog.}, pages 17622--17631, 2022.

\bibitem[Ding et~al.(2020)Ding, Ma, Wang, and Simoncelli]{ding2020image}
Keyan Ding, Kede Ma, Shiqi Wang, and Eero~P Simoncelli.
\newblock Image quality assessment: Unifying structure and texture similarity.
\newblock \emph{IEEE Trans. Pattern Anal. Mach. Intell.}, 44\penalty0 (5):\penalty0 2567--2581, 2020.

\bibitem[Ehret and Facciolo(2019)]{ehret2019study}
Thibaud Ehret and Gabriele Facciolo.
\newblock A study of two cnn demosaicking algorithms.
\newblock \emph{Image Processing On Line}, 9:\penalty0 220--230, 2019.

\bibitem[Ehret et~al.(2019)Ehret, Davy, Arias, and Facciolo]{ehret2019joint}
Thibaud Ehret, Axel Davy, Pablo Arias, and Gabriele Facciolo.
\newblock Joint demosaicking and denoising by fine-tuning of bursts of raw images.
\newblock In \emph{Int. Conf. Comput. Vis.}, pages 8868--8877, 2019.

\bibitem[Eilertsen et~al.(2019)Eilertsen, Mantiuk, and Unger]{eilertsen2019single}
Gabriel Eilertsen, Rafal~K Mantiuk, and Jonas Unger.
\newblock Single-frame regularization for temporally stable cnns.
\newblock In \emph{IEEE Conf. Comput. Vis. Pattern Recog.}, pages 11176--11185, 2019.

\bibitem[Fuoli et~al.(2019)Fuoli, Gu, and Timofte]{fuoli2019efficient}
Dario Fuoli, Shuhang Gu, and Radu Timofte.
\newblock Efficient video super-resolution through recurrent latent space propagation.
\newblock In \emph{Int. Conf. Comput. Vis. Worksh.}, pages 3476--3485. IEEE, 2019.

\bibitem[Gharbi et~al.(2016)Gharbi, Chaurasia, Paris, and Durand]{gharbi2016deep}
Micha{\"e}l Gharbi, Gaurav Chaurasia, Sylvain Paris, and Fr{\'e}do Durand.
\newblock Deep joint demosaicking and denoising.
\newblock \emph{ACM Trans. Graph.}, 35\penalty0 (6):\penalty0 1--12, 2016.

\bibitem[Goodfellow et~al.(2014)Goodfellow, Pouget-Abadie, Mirza, Xu, Warde-Farley, Ozair, Courville, and Bengio]{goodfellow2014generative}
Ian Goodfellow, Jean Pouget-Abadie, Mehdi Mirza, Bing Xu, David Warde-Farley, Sherjil Ozair, Aaron Courville, and Yoshua Bengio.
\newblock Generative adversarial nets.
\newblock \emph{Adv. Neural Inform. Process. Syst.}, 27, 2014.

\bibitem[Guo et~al.(2019)Guo, Yan, Zhang, Zuo, and Zhang]{guo2019toward}
Shi Guo, Zifei Yan, Kai Zhang, Wangmeng Zuo, and Lei Zhang.
\newblock Toward convolutional blind denoising of real photographs.
\newblock In \emph{IEEE Conf. Comput. Vis. Pattern Recog.}, pages 1712--1722, 2019.

\bibitem[Guo et~al.(2021)Guo, Liang, and Zhang]{guo2021joint}
Shi Guo, Zhetong Liang, and Lei Zhang.
\newblock Joint denoising and demosaicking with green channel prior for real-world burst images.
\newblock \emph{IEEE Trans. Image Process.}, 30:\penalty0 6930--6942, 2021.

\bibitem[Guo et~al.(2022)Guo, Yang, Ma, Ren, and Zhang]{guo2022differentiable}
Shi Guo, Xi Yang, Jianqi Ma, Gaofeng Ren, and Lei Zhang.
\newblock A differentiable two-stage alignment scheme for burst image reconstruction with large shift.
\newblock In \emph{IEEE Conf. Comput. Vis. Pattern Recog.}, pages 17472--17481, 2022.

\bibitem[Heide et~al.(2014)Heide, Steinberger, Tsai, Rouf, Pajk, Reddy, Gallo, Liu, Heidrich, Egiazarian, et~al.]{heide2014flexisp}
Felix Heide, Markus Steinberger, Yun-Ta Tsai, Mushfiqur Rouf, Dawid Pajk, Dikpal Reddy, Orazio Gallo, Jing Liu, Wolfgang Heidrich, Karen Egiazarian, et~al.
\newblock Flexisp: A flexible camera image processing framework.
\newblock \emph{ACM Trans. Graph.}, 33\penalty0 (6):\penalty0 1--13, 2014.

\bibitem[Henz et~al.(2018)Henz, Gastal, and Oliveira]{henz2018deep}
Bernardo Henz, Eduardo~SL Gastal, and Manuel~M Oliveira.
\newblock Deep joint design of color filter arrays and demosaicing.
\newblock In \emph{Computer Graphics Forum}, pages 389--399. Wiley Online Library, 2018.

\bibitem[Huang et~al.(2015)Huang, Wang, and Wang]{huang2015bidirectional}
Yan Huang, Wei Wang, and Liang Wang.
\newblock Bidirectional recurrent convolutional networks for multi-frame super-resolution.
\newblock In \emph{Adv. Neural Inform. Process. Syst.}, pages 235--243, 2015.

\bibitem[Huang et~al.(2017)Huang, Wang, and Wang]{huang2017video}
Yan Huang, Wei Wang, and Liang Wang.
\newblock Video super-resolution via bidirectional recurrent convolutional networks.
\newblock \emph{IEEE Trans. Pattern Anal. Mach. Intell.}, 40\penalty0 (4):\penalty0 1015--1028, 2017.

\bibitem[Isobe et~al.(2020{\natexlab{a}})Isobe, Jia, Gu, Li, Wang, and Tian]{isobe2020video2}
Takashi Isobe, Xu Jia, Shuhang Gu, Songjiang Li, Shengjin Wang, and Qi Tian.
\newblock Video super-resolution with recurrent structure-detail network.
\newblock In \emph{Computer Vision--ECCV 2020: 16th European Conference, Glasgow, UK, August 23--28, 2020, Proceedings, Part XII 16}, pages 645--660. Springer, 2020{\natexlab{a}}.

\bibitem[Isobe et~al.(2020{\natexlab{b}})Isobe, Li, Jia, Yuan, Slabaugh, Xu, Li, Wang, and Tian]{isobe2020video}
Takashi Isobe, Songjiang Li, Xu Jia, Shanxin Yuan, Gregory Slabaugh, Chunjing Xu, Ya-Li Li, Shengjin Wang, and Qi Tian.
\newblock Video super-resolution with temporal group attention.
\newblock In \emph{IEEE Conf. Comput. Vis. Pattern Recog.}, pages 8008--8017, 2020{\natexlab{b}}.

\bibitem[Jeelani et~al.(2023)Jeelani, Cheema, Illgner-Fehns, Slusallek, Jaiswal, et~al.]{jeelani2023expanding}
Mehran Jeelani, Noshaba Cheema, Klaus Illgner-Fehns, Philipp Slusallek, Sunil Jaiswal, et~al.
\newblock Expanding synthetic real-world degradations for blind video super resolution.
\newblock In \emph{IEEE Conf. Comput. Vis. Pattern Recog.}, pages 1199--1208, 2023.

\bibitem[Jin et~al.(2020)Jin, Facciolo, and Morel]{jin2020review}
Qiyu Jin, Gabriele Facciolo, and Jean-Michel Morel.
\newblock A review of an old dilemma: Demosaicking first, or denoising first?
\newblock In \emph{IEEE Conf. Comput. Vis. Pattern Recog. Worksh.}, pages 514--515, 2020.

\bibitem[Johnson et~al.(2016)Johnson, Alahi, and Fei-Fei]{johnson2016perceptual}
Justin Johnson, Alexandre Alahi, and Li Fei-Fei.
\newblock Perceptual losses for real-time style transfer and super-resolution.
\newblock In \emph{Eur. Conf. Comput. Vis.}, pages 694--711. Springer, 2016.

\bibitem[Khademi et~al.(2021)Khademi, Rao, Minnerath, Hagen, and Ventura]{khademi2021self}
Wesley Khademi, Sonia Rao, Clare Minnerath, Guy Hagen, and Jonathan Ventura.
\newblock Self-supervised poisson-gaussian denoising.
\newblock In \emph{Proceedings of the IEEE/CVF Winter Conference on Applications of Computer Vision}, pages 2131--2139, 2021.

\bibitem[Kingma and Ba(2014)]{kingma2014adam}
Diederik~P Kingma and Jimmy Ba.
\newblock Adam: A method for stochastic optimization.
\newblock \emph{arXiv preprint arXiv:1412.6980}, 2014.

\bibitem[Kokkinos and Lefkimmiatis(2019)]{kokkinos2019iterative}
Filippos Kokkinos and Stamatios Lefkimmiatis.
\newblock Iterative joint image demosaicking and denoising using a residual denoising network.
\newblock \emph{IEEE Trans. Image Process.}, 28\penalty0 (8):\penalty0 4177--4188, 2019.

\bibitem[Lai et~al.(2018)Lai, Huang, Wang, Shechtman, Yumer, and Yang]{lai2018learning}
Wei-Sheng Lai, Jia-Bin Huang, Oliver Wang, Eli Shechtman, Ersin Yumer, and Ming-Hsuan Yang.
\newblock Learning blind video temporal consistency.
\newblock In \emph{Eur. Conf. Comput. Vis.}, pages 170--185, 2018.

\bibitem[Lei and Chen(2019)]{lei2019fully}
Chenyang Lei and Qifeng Chen.
\newblock Fully automatic video colorization with self-regularization and diversity.
\newblock In \emph{IEEE Conf. Comput. Vis. Pattern Recog.}, pages 3753--3761, 2019.

\bibitem[Liu et~al.(2018)Liu, Wen, Fan, Loy, and Huang]{liu2018non}
Ding Liu, Bihan Wen, Yuchen Fan, Chen~Change Loy, and Thomas~S Huang.
\newblock Non-local recurrent network for image restoration.
\newblock \emph{Adv. Neural Inform. Process. Syst.}, 31, 2018.

\bibitem[Liu et~al.(2020{\natexlab{a}})Liu, Jia, Liu, and Tian]{liu2020joint}
Lin Liu, Xu Jia, Jianzhuang Liu, and Qi Tian.
\newblock Joint demosaicing and denoising with self guidance.
\newblock In \emph{IEEE Conf. Comput. Vis. Pattern Recog.}, pages 2240--2249, 2020{\natexlab{a}}.

\bibitem[Liu et~al.(2020{\natexlab{b}})Liu, Chen, Xun, Zhao, and Chang]{liu2020new}
Shumin Liu, Jiajia Chen, Yuan Xun, Xiaojin Zhao, and Chip-Hong Chang.
\newblock A new polarization image demosaicking algorithm by exploiting inter-channel correlations with guided filtering.
\newblock \emph{IEEE Trans. Image Process.}, 29:\penalty0 7076--7089, 2020{\natexlab{b}}.

\bibitem[Loshchilov and Hutter(2016)]{loshchilov2016sgdr}
Ilya Loshchilov and Frank Hutter.
\newblock Sgdr: Stochastic gradient descent with warm restarts.
\newblock \emph{arXiv preprint arXiv:1608.03983}, 2016.

\bibitem[Luisier et~al.(2010)Luisier, Blu, and Unser]{luisier2010image}
Florian Luisier, Thierry Blu, and Michael Unser.
\newblock Image denoising in mixed poisson--gaussian noise.
\newblock \emph{IEEE Trans. Image Process.}, 20\penalty0 (3):\penalty0 696--708, 2010.

\bibitem[Mildenhall et~al.(2018)Mildenhall, Barron, Chen, Sharlet, Ng, and Carroll]{mildenhall2018burst}
Ben Mildenhall, Jonathan~T Barron, Jiawen Chen, Dillon Sharlet, Ren Ng, and Robert Carroll.
\newblock Burst denoising with kernel prediction networks.
\newblock In \emph{IEEE Conf. Comput. Vis. Pattern Recog.}, pages 2502--2510, 2018.

\bibitem[Nah et~al.(2019)Nah, Baik, Hong, Moon, Son, Timofte, and Mu~Lee]{nah2019ntire}
Seungjun Nah, Sungyong Baik, Seokil Hong, Gyeongsik Moon, Sanghyun Son, Radu Timofte, and Kyoung Mu~Lee.
\newblock Ntire 2019 challenge on video deblurring and super-resolution: Dataset and study.
\newblock In \emph{IEEE Conf. Comput. Vis. Pattern Recog. Worksh.}, pages 0--0, 2019.

\bibitem[Park et~al.(2019)Park, Woo, Kim, Cho, and Kweon]{park2019preserving}
Kwanyong Park, Sanghyun Woo, Dahun Kim, Donghyeon Cho, and In~So Kweon.
\newblock Preserving semantic and temporal consistency for unpaired video-to-video translation.
\newblock In \emph{ACM Int. Conf. Multimedia}, pages 1248--1257, 2019.

\bibitem[pixabey(2020)]{pixabeydataset}
pixabey.
\newblock pixabey website.
\newblock \url{https://www.pixabay.com/}, 2020.

\bibitem[Pl{\"o}tz and Roth(2018)]{plotz2018neural}
Tobias Pl{\"o}tz and Stefan Roth.
\newblock Neural nearest neighbors networks.
\newblock \emph{Adv. Neural Inform. Process. Syst.}, 31, 2018.

\bibitem[Qian et~al.(2019)Qian, Gu, Ren, Dong, Zhao, and Lin]{qian2019trinity}
Guocheng Qian, Jinjin Gu, Jimmy~S Ren, Chao Dong, Furong Zhao, and Juan Lin.
\newblock Trinity of pixel enhancement: a joint solution for demosaicking, denoising and super-resolution.
\newblock \emph{arXiv preprint arXiv:1905.02538}, 2019.

\bibitem[Ranjan and Black(2017)]{ranjan2017optical}
Anurag Ranjan and Michael~J Black.
\newblock Optical flow estimation using a spatial pyramid network.
\newblock In \emph{IEEE Conf. Comput. Vis. Pattern Recog.}, pages 4161--4170, 2017.

\bibitem[Sajjadi et~al.(2018)Sajjadi, Vemulapalli, and Brown]{sajjadi2018frame}
Mehdi~SM Sajjadi, Raviteja Vemulapalli, and Matthew Brown.
\newblock Frame-recurrent video super-resolution.
\newblock In \emph{IEEE Conf. Comput. Vis. Pattern Recog.}, pages 6626--6634, 2018.

\bibitem[Song et~al.(2022)Song, Zhang, and Ayd{\i}n]{song2022tempformer}
Mingyang Song, Yang Zhang, and Tun{\c{c}}~O Ayd{\i}n.
\newblock Tempformer: Temporally consistent transformer for video denoising.
\newblock In \emph{Eur. Conf. Comput. Vis.}, pages 481--496. Springer, 2022.

\bibitem[Wang et~al.(2018)Wang, Liu, Zhu, Liu, Tao, Kautz, and Catanzaro]{wang2018video}
Ting-Chun Wang, Ming-Yu Liu, Jun-Yan Zhu, Guilin Liu, Andrew Tao, Jan Kautz, and Bryan Catanzaro.
\newblock Video-to-video synthesis.
\newblock \emph{arXiv preprint arXiv:1808.06601}, 2018.

\bibitem[Wang et~al.(2020)Wang, Xu, Zhang, Wang, and Liu]{wang2020consistent}
Wenjing Wang, Jizheng Xu, Li Zhang, Yue Wang, and Jiaying Liu.
\newblock Consistent video style transfer via compound regularization.
\newblock In \emph{AAAI}, pages 12233--12240, 2020.

\bibitem[Wang et~al.(2019)Wang, Chan, Yu, Dong, and Change~Loy]{wang2019edvr}
Xintao Wang, Kelvin~CK Chan, Ke Yu, Chao Dong, and Chen Change~Loy.
\newblock Edvr: Video restoration with enhanced deformable convolutional networks.
\newblock In \emph{IEEE Conf. Comput. Vis. Pattern Recog. Worksh.}, pages 0--0, 2019.

\bibitem[Wu et~al.(2022)Wu, Wang, Li, and Shan]{wu2022animesr}
Yanze Wu, Xintao Wang, Gen Li, and Ying Shan.
\newblock Animesr: learning real-world super-resolution models for animation videos.
\newblock \emph{Adv. Neural Inform. Process. Syst.}, 35:\penalty0 11241--11252, 2022.

\bibitem[Xie et~al.(2023)Xie, Wang, Shi, Gu, Dong, and Shan]{xie2023mitigating}
Liangbin Xie, Xintao Wang, Shuwei Shi, Jinjin Gu, Chao Dong, and Ying Shan.
\newblock Mitigating artifacts in real-world video super-resolution models.
\newblock In \emph{AAAI}, pages 2956--2964, 2023.

\bibitem[Xu et~al.(2019)Xu, Li, and Sun]{xu2019learning}
Xiangyu Xu, Muchen Li, and Wenxiu Sun.
\newblock Learning deformable kernels for image and video denoising.
\newblock \emph{arXiv preprint arXiv:1904.06903}, 2019.

\bibitem[Yan and Ouyang(2019)]{yan2019cross}
Niu Yan and Jihong Ouyang.
\newblock Cross-channel correlation preserved three-stream lightweight cnns for demosaicking.
\newblock \emph{arXiv preprint arXiv:1906.09884}, 2019.

\bibitem[Yang and Wang(2019)]{yang2019efficient}
Bin Yang and Dongsheng Wang.
\newblock An efficient adaptive interpolation for bayer cfa demosaicking.
\newblock \emph{Sensing and Imaging}, 20\penalty0 (1):\penalty0 37, 2019.

\bibitem[Yang et~al.(2021)Yang, Xiang, Zeng, and Zhang]{yang2021real}
Xi Yang, Wangmeng Xiang, Hui Zeng, and Lei Zhang.
\newblock Real-world video super-resolution: A benchmark dataset and a decomposition based learning scheme.
\newblock In \emph{Int. Conf. Comput. Vis.}, pages 4781--4790, 2021.

\bibitem[Yue et~al.(2020)Yue, Cao, Liao, Chu, and Yang]{yue2020supervised}
Huanjing Yue, Cong Cao, Lei Liao, Ronghe Chu, and Jingyu Yang.
\newblock Supervised raw video denoising with a benchmark dataset on dynamic scenes.
\newblock In \emph{IEEE Conf. Comput. Vis. Pattern Recog.}, pages 2301--2310, 2020.

\bibitem[Zhang et~al.(2021)Zhang, Li, You, and Fu]{zhang2021learning}
Fan Zhang, Yu Li, Shaodi You, and Ying Fu.
\newblock Learning temporal consistency for low light video enhancement from single images.
\newblock In \emph{IEEE Conf. Comput. Vis. Pattern Recog.}, pages 4967--4976, 2021.

\bibitem[Zhang et~al.(2017)Zhang, Zuo, Chen, Meng, and Zhang]{zhang2017beyond}
Kai Zhang, Wangmeng Zuo, Yunjin Chen, Deyu Meng, and Lei Zhang.
\newblock Beyond a gaussian denoiser: Residual learning of deep cnn for image denoising.
\newblock \emph{IEEE Trans. Image Process.}, 26\penalty0 (7):\penalty0 3142--3155, 2017.

\bibitem[Zhang et~al.(2018{\natexlab{a}})Zhang, Zuo, and Zhang]{zhang2018ffdnet}
Kai Zhang, Wangmeng Zuo, and Lei Zhang.
\newblock Ffdnet: Toward a fast and flexible solution for cnn-based image denoising.
\newblock \emph{IEEE Trans. Image Process.}, 27\penalty0 (9):\penalty0 4608--4622, 2018{\natexlab{a}}.

\bibitem[Zhang et~al.(2018{\natexlab{b}})Zhang, Isola, Efros, Shechtman, and Wang]{zhang2018unreasonable}
Richard Zhang, Phillip Isola, Alexei~A Efros, Eli Shechtman, and Oliver Wang.
\newblock The unreasonable effectiveness of deep features as a perceptual metric.
\newblock In \emph{IEEE Conf. Comput. Vis. Pattern Recog.}, pages 586--595, 2018{\natexlab{b}}.

\end{thebibliography}
}


\end{document}